\def\year{2021}\relax
\documentclass[letterpaper]{article} % DO NOT CHANGE THIS
\usepackage{aaai21}  % DO NOT CHANGE THIS
\usepackage{times}  % DO NOT CHANGE THIS
\usepackage{helvet} % DO NOT CHANGE THIS
\usepackage{courier}  % DO NOT CHANGE THIS
\usepackage[hyphens]{url}  % DO NOT CHANGE THIS
\usepackage{graphicx} % DO NOT CHANGE THIS
\usepackage{natbib}  % DO NOT CHANGE THIS AND DO NOT ADD ANY OPTIONS TO IT
\usepackage{caption} % DO NOT CHANGE THIS AND DO NOT ADD ANY OPTIONS TO IT
\usepackage{amsmath}
\usepackage{amssymb}
\usepackage{amsthm}
\usepackage{tikz}
\usepackage{algorithm}
\usepackage{algorithmic}
\newlength\myindent
\usepackage{booktabs}
\usepackage{array}
\usepackage{multirow}
\usepackage{textcomp}

\begin{document}

\title{Proximal Policy Gradient: PPO with Policy Gradient}
\author{

    %Authors
    % All authors must be in the same font size and format.
    Ju-Seung Byun\textsuperscript{\rm 1},
    Byungmoon Kim\textsuperscript{\rm 2},
    Huamin Wang\textsuperscript{\rm 1}
    \\
}
\affiliations{
    \textsuperscript{\rm 1}Ohio State University,
    \textsuperscript{\rm 2}Adobe Research\\
    byun.83@osu.edu, bmkim@adobe.com, whmin@cse.ohio-state.edu
}

\maketitle

\begin{abstract}
In this paper, we propose a new algorithm PPG (Proximal Policy Gradient), which is close to both VPG (vanilla policy gradient) and PPO (proximal policy optimization). The PPG objective is a partial variation of the VPG objective and the gradient of the PPG objective is exactly same as the gradient of the VPG objective. To increase the number of policy update iterations, we introduce the advantage-policy plane and design a new clipping strategy. We perform experiments in OpenAI Gym and Bullet robotics environments for ten random seeds. The performance of PPG is comparable to PPO, and the entropy decays slower than PPG. Thus we show that performance similar to PPO can be obtained by using the gradient formula from the original policy gradient theorem.
\end{abstract}

\section{Introduction}
Policy gradient algorithms introduced in \cite{williams1992simple, NIPS1999_1713, sutton2000policy} compute the gradient of the expected return with respect to policy parameters. An action-independent term, typically the state value function, can be subtracted from this gradient. Then, the policy gradient is in the form of gradient of log probability multiplied by the advantage function. The resulting policy gradient algorithm respectively boosts (reduces), the probability of advantageous (disadvantageous) actions. This policy gradient algorithm is too slow to be used as a modern day baseline. Therefore, a reasonable choice of recent advances are applied to the classic policy gradient theorem based methods. In \cite{SpinningUp2018}, the generalized advantage estimate (GAE) \cite{schulman2015high} and the advantage normalization are applied. The resulting algorithm is called the Vanilla Policy Gradient (VPG) \cite{SpinningUp2018, baselines}.

Recently, the trust region policy optimization (TRPO) \cite{pmlr-v37-schulman15} and the proximal policy optimization (PPO) \cite{DBLP:journals/corr/SchulmanWDRK17} present significant performance improvement over VPG by using the proposed surrogate objective function. In particular, PPO is very popular due to the learning stability and implementation simplicity. PPO provides a novel solution to the problem of increasing the number of policy updates by introducing the clipping idea. In practice, the number of iterations cannot be indefinitely large. Instead, the iteration continues until a target KL divergence is reached. Thus a much larger number of policy update iterations per each roll out is possible, greatly increasing the sample efficiency while maintaining the steady improvement on the average return. This is the baseline PPO algorithm we use to perform experiments. Note that there are variety of variations of PPO \cite{DBLP:journals/corr/SchulmanWDRK17} for various enhancements.

We explore a slight modification to both VPG and PPO, devise an objective that appears to be in between, and study the performance. The objective without clipping has the gradient exactly the same as VPG's gradient. This objective can be considered as a partial variation of the VPG objective, i.e., PPG optimizes the changes of the VPG objective. In contrast, PPO optimizes the expected importance-weighted advantage, and hence, the gradient of the PPO objective is different from VPG, which is based on the policy gradient theorem \cite{NIPS1999_1713}. To achieve performance similar to PPO, we introduce a new clipping algorithm designed for the PPG objective. The clipping is applied to samples that have their probabilities adjusted enough, so that these samples are excluded from the training batch.

We perform experiments in the popular continuous control environments in OpenAI Gym \cite{DBLP:journals/corr/BrockmanCPSSTZ16} with the MuJoCo simulator: Ant, HalfCheetah, Hopper, and Walker2d, and also in the corresponding Bullet environments: AntBulletEnv, HalfCheetahBulletEnv, HopperBulletEnv, and Walker2DBulletEnv. We found that PPG and PPO are very similar with a few cases where PPG appears to have marginally higher average returns. The entropy tends to decay slower in PPG than in PPO. Therefore, PPG solutions appear to be more stochastic than PPO solutions. Thus, we show that policy gradient can be used for the PPO algorithm.

\section{Related Work}
Model free reinforcement learning algorithms have evolved from the classic dynamic programming, policy and value iterations \cite{SuttonBarto:2018} to Q-learning \cite{watkins1989learning} and direct policy search methods using neural networks for policy and value functions \cite{mnih2015humanlevel,mnih2016asynchronous,Marcin2018Hindsight,lillicrap2015continuous,Fujimoto2018,haarnoja2018soft}. 

On-policy algorithms are popular for their stability, i.e., applicability to a wide set of problems. One of the early on-policy policy-gradient based methods is REINFORCE \cite{williams1992simple} that directly used Monte-Carlo returns. Instead, value estimate was used in the Actor-Critic method \cite{Sutton1999AC}. In modern days, value estimates are frequently formulated as advantages. To reduce the noise in advantage, the generalized advantage estimation method has been proposed in \cite{schulman2015high}.

The policy-gradient theorem \cite{NIPS1999_1713} is used to compute the gradient of expected return in terms of policy parameters. This gradient is formulated as an objective function, which is handed over to an optimizer with an auto-derivative support. A reasonable implementation of this algorithm is called VPG  \cite{SpinningUp2018, baselines}. On the other hand, TRPO \cite{pmlr-v37-schulman15} and PPO \cite{DBLP:journals/corr/SchulmanWDRK17} use the  surrogate objective. TRPO uses natural policy gradient with KL divergence constraints while PPO uses a simple clipping mechanism. Despite the simplicity, PPO often outperforms TRPO.

The performance of PPO can be improved by a variety of secondary optimization techniques. For example, value function clipping and reward scaling have been studied in \cite{Engstrom2020Implementation}, where these secondary optimizations are shown to be crucial for PPO's performance. Since the success of model-free algorithms depends on its exploration and PPO tends to not sufficiently explore, \citeauthor{wang2019trust} \shortcite{wang2019trust} proposed to adjust the clipping range within the trust region to encourage the exploration. The performance of PPO is also enormously influenced by its entropy. \citeauthor{ahmed2019understanding} \shortcite{ahmed2019understanding} and \citeauthor{liu2019regularization} \shortcite{liu2019regularization} present higher entropy can be utilized as a regularizer that produces smoother training results. 

In this paper, we choose a PPO implementation with clipping and KL divergence break, and then study whether the original policy gradient theorem can be used instead of the surrogate objective, and still performance similar to PPO can be obtained.

\section{Policy Optimization Objectives}

In this section, we first define the objectives of VPG, PPO, and PPG, and provide the detailed discussions in the later sections.

\subsection{VPG: Vanilla Policy Gradient}
Policy Gradient (PG) methods, originally developed in \cite{williams1992simple, sutton2000policy}, are to update a parameterized policy by ascending along the stochastic gradient of the expected return. The gradient of PG has the form 
\begin{align}
    \nabla_{\theta} J(\pi_\theta) = \mathbb{E}_{\pi_\theta}\bigg[ \nabla_{\theta}\log\pi_\theta(a_{t} | s_{t})A^{\pi}(s_{t},a_{t})\bigg],
\end{align}
where $\pi_\theta$ is a stochastic policy with the parameter $\theta$. $A^{\pi}$ is commonly referred to as the advantage function $A^{\pi}(s_{t}, a_{t}) = Q^{\pi}(s_t,a_t) - V^{\pi}(s_t)$ where $Q^{\pi}(s_t, a_t)$ is a state-action value function and  $V^{\pi}(s_{t})$ is a state value function. Since this advantage $A^\pi_t$ is noisy, the generalized advantage estimation \cite{schulman2015high} is applied to reduce the variance. We denote the resulting advantage as $A_{t}^{GAE}$. Finally, we normalize $A_{t}^{GAE}$ and compute the normalized advantage $\hat A_t$ such that the mean is zero and the standard deviation is 1. We exploit this $\hat A_t$ to define the objective of VPG, denoted as $L^{\scalebox{0.55}{$VPG$}}$.
\begin{align}
    L^{\scalebox{0.55}{$VPG$}}  = \frac{1}{\hat N}\sum_{i,t=1}^{N,T_i} \log\pi_{\theta}(a_{i,t}|s_{i,t}) \hat A_t.
\label{eq:Loss_VPG}\end{align}
where $N$ is the number of episodes in the roll out, $T_i$ is the time steps in episode $i$, and $\hat N = \sum T_i$ is the size of the roll out.

\subsection{PPO: Proximal Policy Optimization}
Proximal Policy Optimization (PPO) \cite{DBLP:journals/corr/SchulmanWDRK17} and TRPO \cite{wang2019trust} use the surrogate objective \cite{zhang2018natural, pmlr-v37-schulman15}. 
\begin{equation}\begin{aligned}\begin{split}
    & \underset{\theta}{\mathrm{maximize}}\; \mathbb{E}_{t}\bigg[\frac{\pi_{\theta}(a_{t} | s_{t})}{\pi_{\theta_{old}}(a_{t}|s_{t})}\hat A_t  \bigg] \\
    & \text{subject to}\; \mathbb{E}_{t}\bigg[ D_{KL}\left(\pi_{\theta}(\cdot|s_{t}) || \pi_{\theta_{old}}(\cdot|s_{t})\right)\bigg] \leq \delta
\end{split}\end{aligned}\label{eq:Surrogate}\end{equation}
To avoid the high computational cost and implementation complexity of TRPO, PPO proposes simple techniques: clipped surrogate objective or adaptive KL penalty \cite{DBLP:journals/corr/SchulmanWDRK17}.

Among many variations of PPO, we select an OpenAI's implementation found in \cite{SpinningUp2018}. The loss objective of PPO using the clipped surrogate objective method is $L^{\scalebox{0.55}{$PPO$}}$ in \eqref{eq:Loss_PPO}. We also provide the objective without clipping $L^{\scalebox{0.55}{$PPO$}}_{\scalebox{0.55}{$NCLIP$}}$.
\begin{align}\begin{split}
    &L^{\scalebox{0.55}{$PPO$}}_{\scalebox{0.55}{$NCLIP$}}  = \frac{1}{\hat N}\sum_{i, t=1}^{N,T_i} r_{i,t}(\theta) \hat A_t  \\
    &L^{\scalebox{0.55}{$PPO$}}  = \frac{1}{\hat N}\sum_{i, t=1}^{N,T_i} 
    \mathrm{min} (r_{i,t}(\theta), \mathrm{clip}  ( r_{i,t}(\theta), 1 \pm \epsilon) ) \hat A_t  \\ 
    &\qquad\; \text{where} \;r_{i,t}(\theta) = \pi_{\theta}(a_{i,t} | s_{i,t})\; / \; \pi_{\theta_{old}}(a_{i,t}|s_{i,t})
\label{eq:Loss_PPO}\end{split}\end{align}

\begin{figure*}[t!]
\centering
\includegraphics[width=1.9\columnwidth]{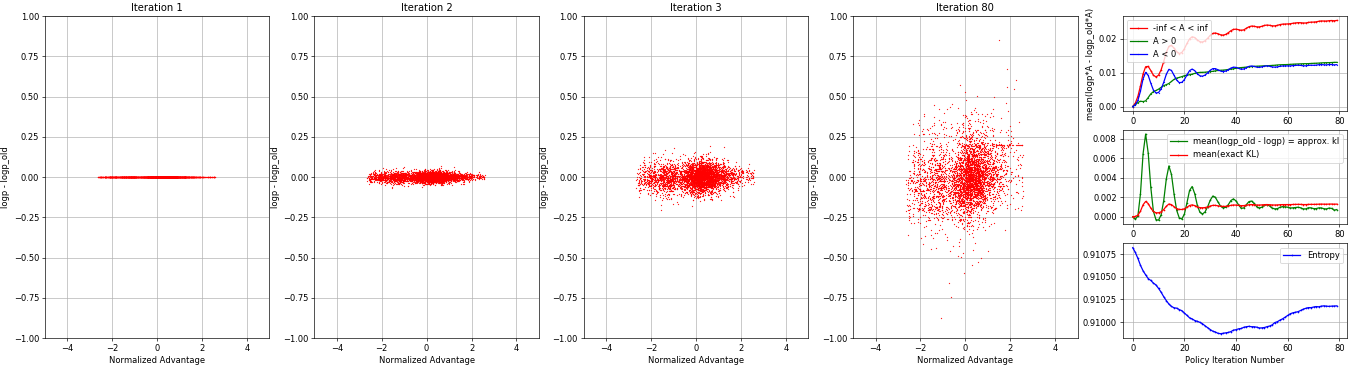}
\caption{Typical PPG policy iterations from AntBulletEnv training with roll out size of 4000. At the beginning of the iteration, all samples are in the $x$ axis as $\log\pi_i - \log\pi_{i_{\mbox{\scriptsize old}}} = 0$. As iteration continues, policy is updated, and $\log\pi_i - \log\pi_{i_{\mbox{\scriptsize old}}}$ changes. In the far right, we show the objective changes in top (green for positive advantages, blue for negative advantages, and red for all), KL divergences in the middle (green for approximate KL and red for the average exact KL), and entropy changes in the bottom.}
\label{fig_ppg_po}
\end{figure*}

\subsection{PPG: Proximal Policy Gradient}
PPG, which we propose in this paper, is a simple variant of VPG, or a simple variant of PPO. We design the objective so that its gradient is the same as the gradient of $L^{\scalebox{0.55}{$VPG$}}$, while a clipping method similar to PPO is applicable. The resulting form has the log probability difference between the current policy and the old policy, $d_{i,t}(\theta) = \log \pi_{\theta}(a_{i,t}|s_{i,t}) - \log \pi_{\theta_{old}}(a_{i,t}|s_{i,t})$. This can also be interpreted as the application of the logarithmic operation to the surrogate objective as $d_{i,t} = \log r_{i,t}$. The objectives without clipping $L^{\scalebox{0.55}{$PPG$}}_{\scalebox{0.55}{$NCLIP$}}$ and with clipping $L^{\scalebox{0.55}{$PPG$}}$ are 
\begin{align}\begin{split}
    &L^{\scalebox{0.55}{$PPG$}}_{\scalebox{0.55}{$NCLIP$}} = \frac{1}{\hat N}\sum_{i, t=1}^{N,T_i} d_{i,t}(\theta) \hat A_t,
    \\
    &L^{\scalebox{0.55}{$PPG$}} = \frac{1}{\hat N}\sum_{i, t=1}^{N,T_i} \hat A_t \; \delta_{i,t}(\theta),\\
    &\quad \delta_{i,t}(\theta) =\left\{\begin{array}{ll}
            \min(d_{i,t}(\theta) , u_b),\; \hat A_t \geq 0\\
            \max(d_{i,t}(\theta) , l_b),\; \hat A_t < 0
        \end{array}
        \right.,
    \\
    &\quad d_{i,t}(\theta) = \log \pi_{\theta}(a_{i,t}|s_{i,t}) - \log \pi_{\theta_{old}}(a_{i,t}|s_{i,t}), \\
    &\quad u_b \; \text{is the clipping upper bound constant,} \\
    &\quad l_b \; \text{is the clipping lower bound constant.} \\
\end{split}\label{eq:Loss_PPG}
\end{align}
Note that this clipping depends on the sign of the advantage $\hat A_t$. For the design details of $L^{\scalebox{0.55}{$PPG$}}$, we defer the discussion to the next section.

Since $\pi_{\theta_{old}}$ is a constant, the derivative of the $L^{\scalebox{0.55}{$PPG$}}_{\scalebox{0.55}{$NCLIP$}}$ is exactly equal to $L^{\scalebox{0.55}{$VPG$}}$, which is based on the gradient from the policy gradient theorem. 
Also note that the gradient of $L^{\scalebox{0.55}{$PPO$}}_{\scalebox{0.55}{$NCLIP$}}$ is different from $L^{\scalebox{0.55}{$VPG$}}$.
\begin{align}\begin{split}
    \nabla_{\theta} L^{\scalebox{0.55}{$PPG$}}_{\scalebox{0.55}{$NCLIP$}}
       &= \nabla_{\theta} L^{\scalebox{0.55}{$VPG$}} =  \frac{1}{\hat N}\sum_{i, t=1}^{N,T_i}
       \nabla_{\theta} \log \pi_{\theta}(a_{i,t}|s_{i,t})\hat A_t \\
       &\neq  \nabla_{\theta} L^{\scalebox{0.55}{$PPO$}}_{\scalebox{0.55}{$NCLIP$}} = \frac{1}{\hat N}\sum_{i, t=1}^{N,T_i} \frac{\nabla_{\theta} \pi_{\theta}(a_{i,t}|s_{i, t})}{\pi_{\theta_{old}}(a_{i,t}|s_{i,t})}\hat A_t
%\label{eq:Derivatives}
\nonumber
\end{split}\end{align}
The samples clipped by \eqref{eq:Loss_PPG} do not contribute to the gradient of $L^{\scalebox{0.55}{$PPG$}}$, and hence, for unclipped samples, $\nabla_\theta L^{\scalebox{0.55}{$PPG$}}=\nabla_{\theta} L^{\scalebox{0.55}{$PPG$}}_{\scalebox{0.55}{$NCLIP$}} = \nabla_{\theta} L^{\scalebox{0.55}{$VPG$}}$.

\section{PPG Clipping in Advantage Policy Plane}
\begin{figure}[h]
\centering
\includegraphics[width=1.0\columnwidth]{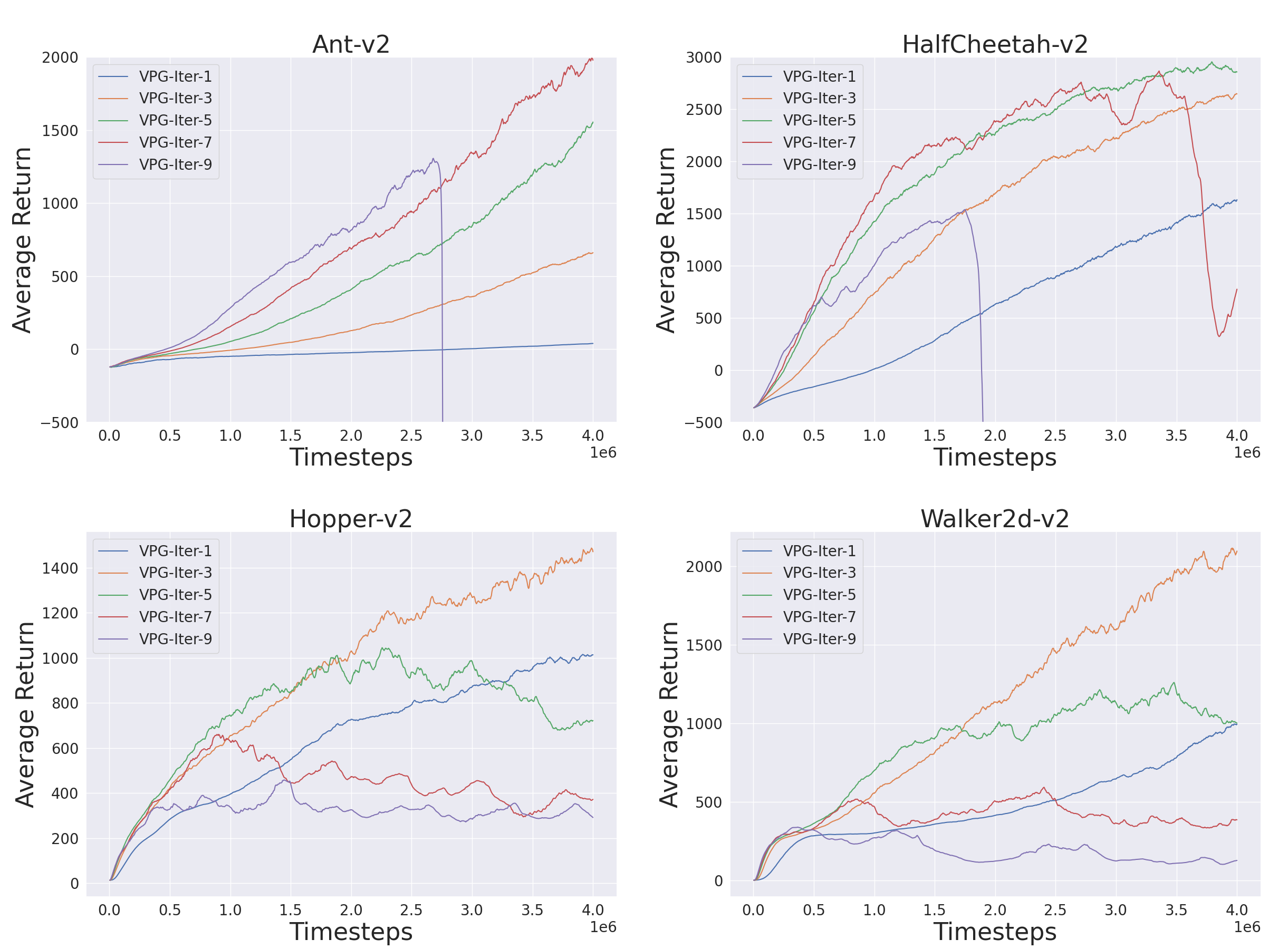}
\caption{Results from multiple VPG policy iterations are exactly the same as PPG iterations without clipping, i.e., optimizing with $L^{\scalebox{0.55}{$PPG$}}_{\scalebox{0.55}{$NCLIP$}}$. Simply increasing the number of VPG policy iterations can result in eventual failures. Even for iterations that do not demonstrate a failure in this experiment, a failure can happen later if the training is continued.}
\label{fig:Iteration}
\end{figure}

When training a policy using VPG, it is difficult to match the performance of PPO simply by increasing the number of iteration as shown in Fig. \ref{fig:Iteration}. In order to achieve similar performance of PPO, a clipping mechanism similar to PPO is needed. In this section, we develop this clipping formula $L^{\scalebox{0.55}{$PPG$}}$ in \eqref{eq:Loss_PPG}, which we call PPG clipping.

\begin{figure*}[t]
\centering
\includegraphics[width=0.5\columnwidth]{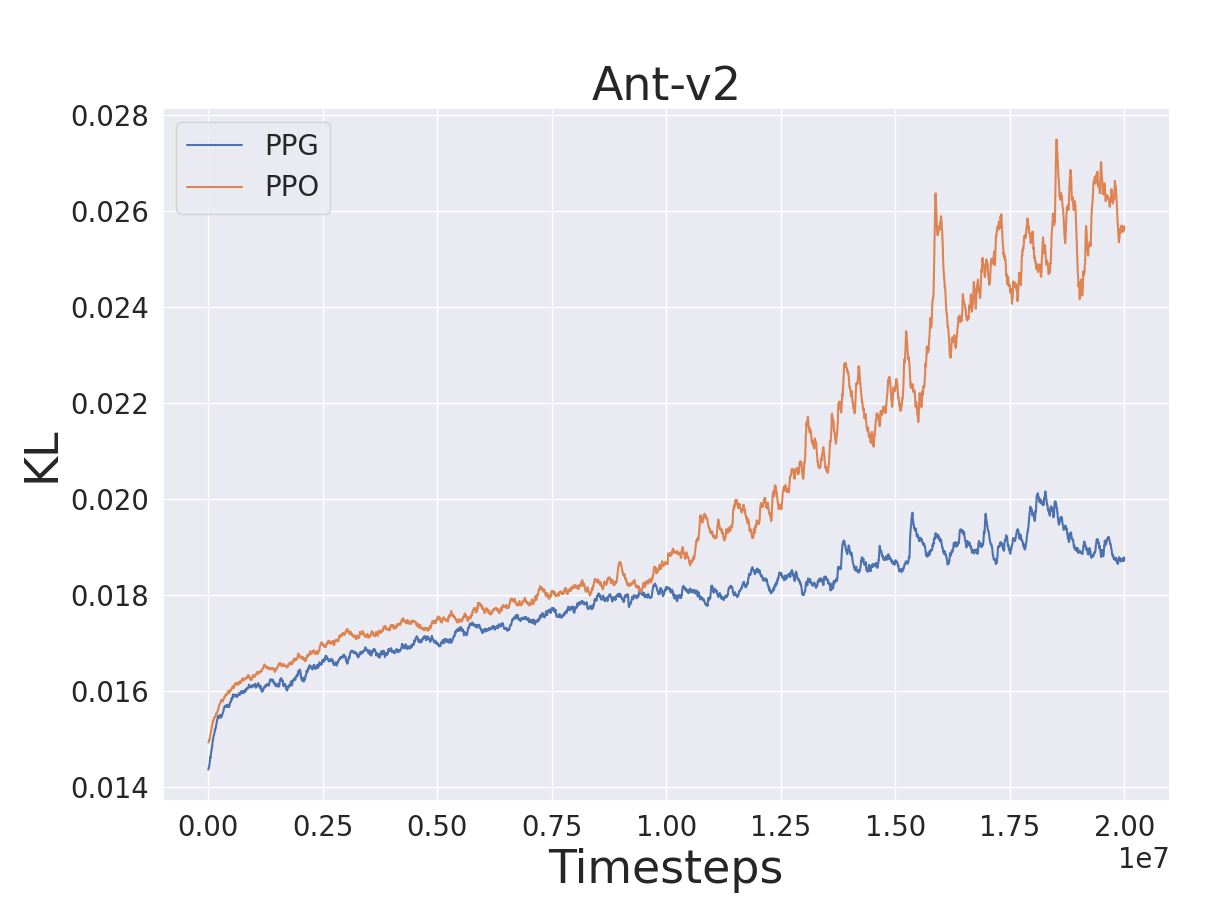} 
\includegraphics[width=0.5\columnwidth]{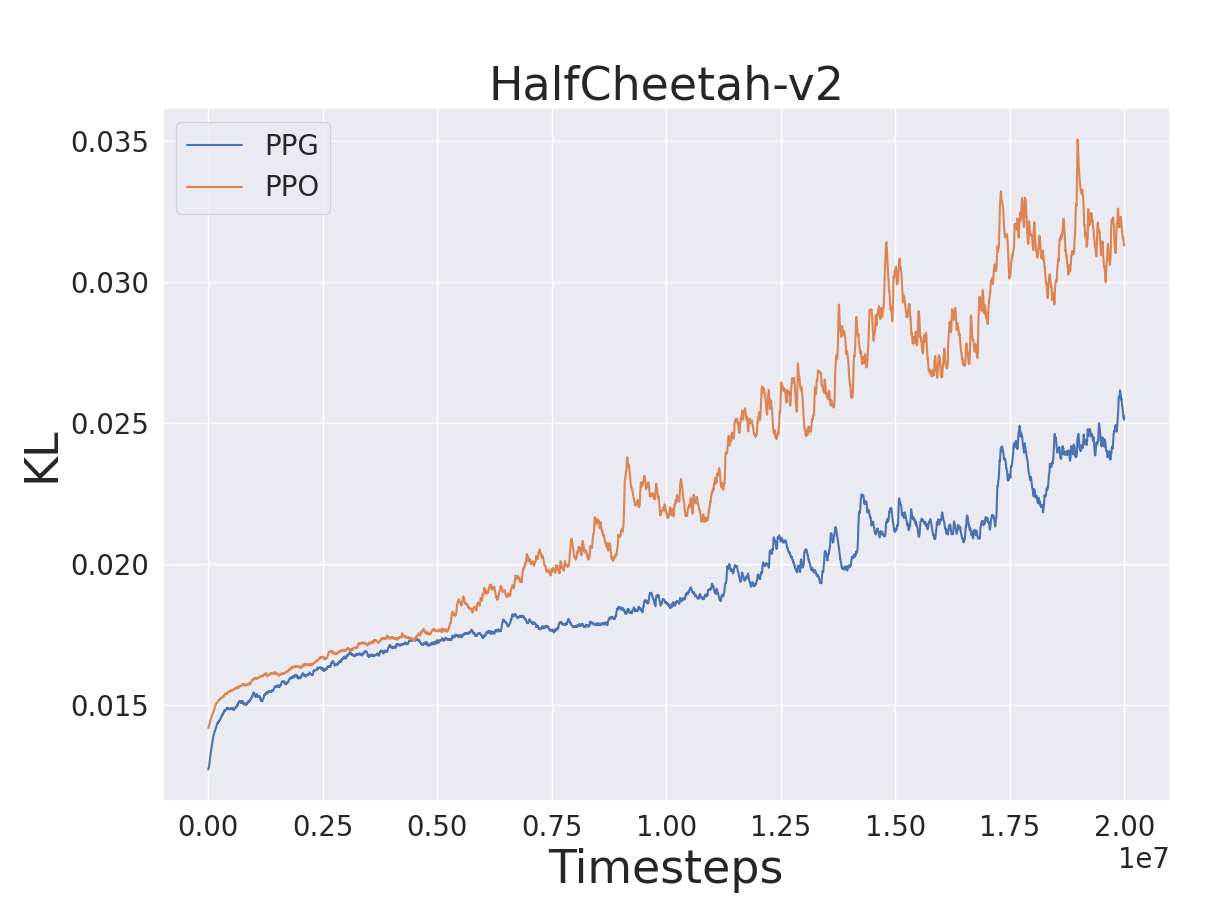}
\includegraphics[width=0.5\columnwidth]{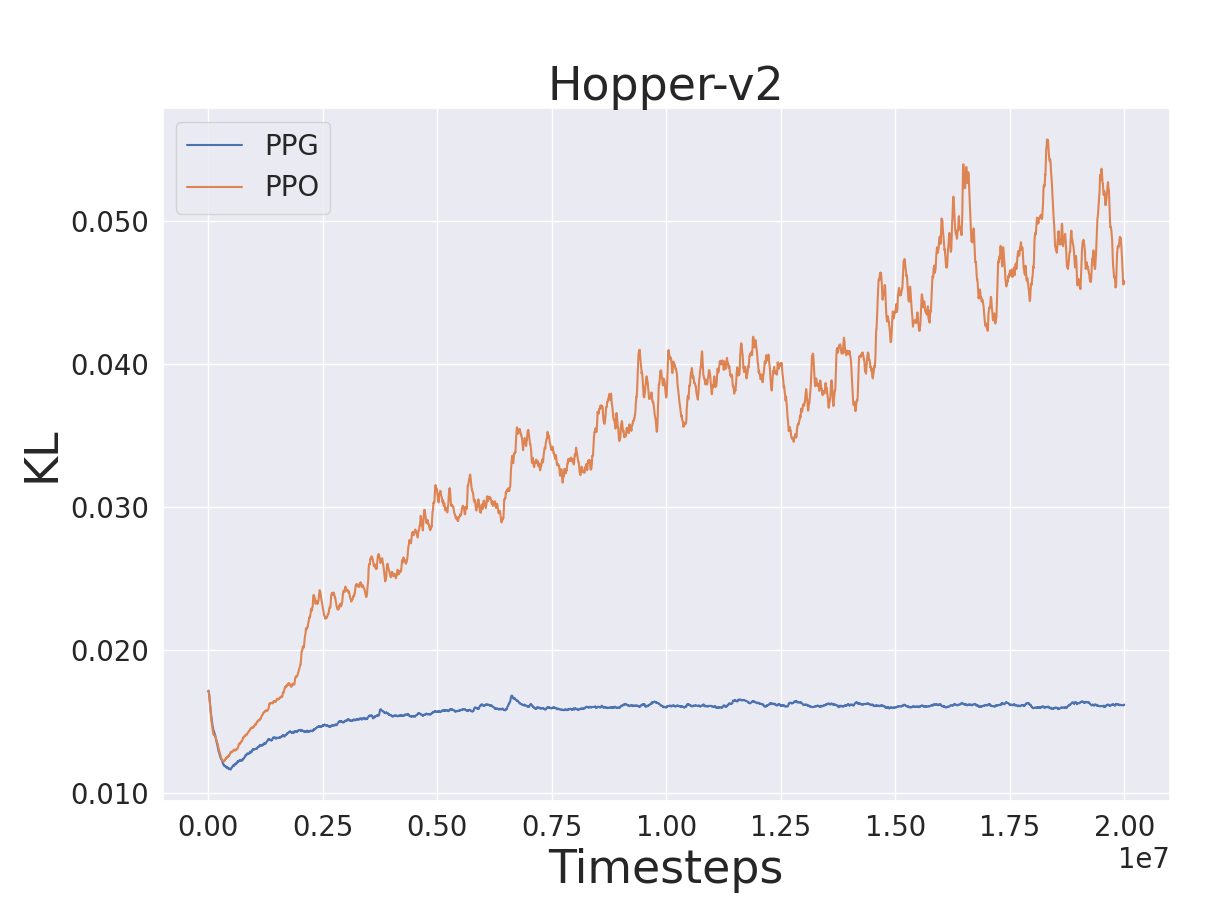}
\includegraphics[width=0.5\columnwidth]{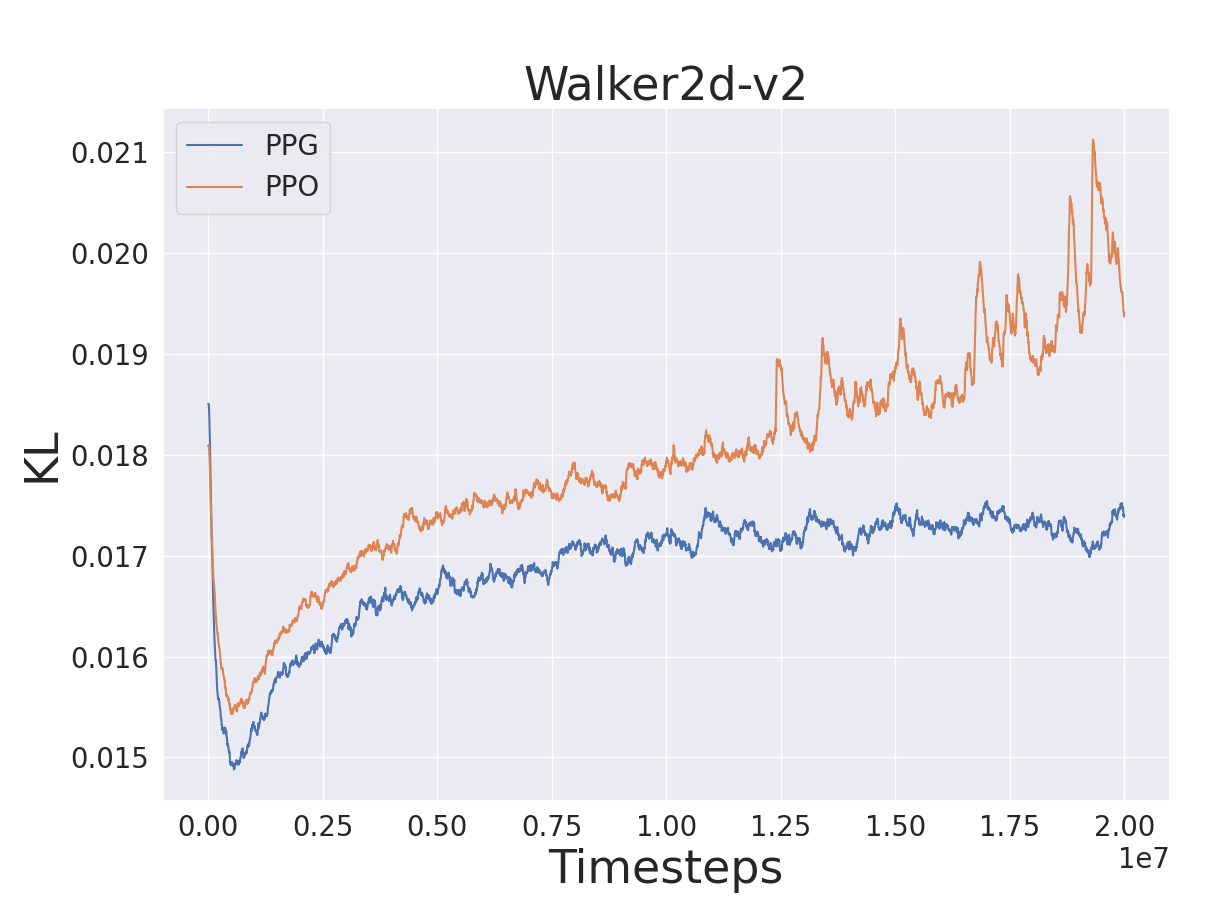} \\
\includegraphics[width=0.5\columnwidth]{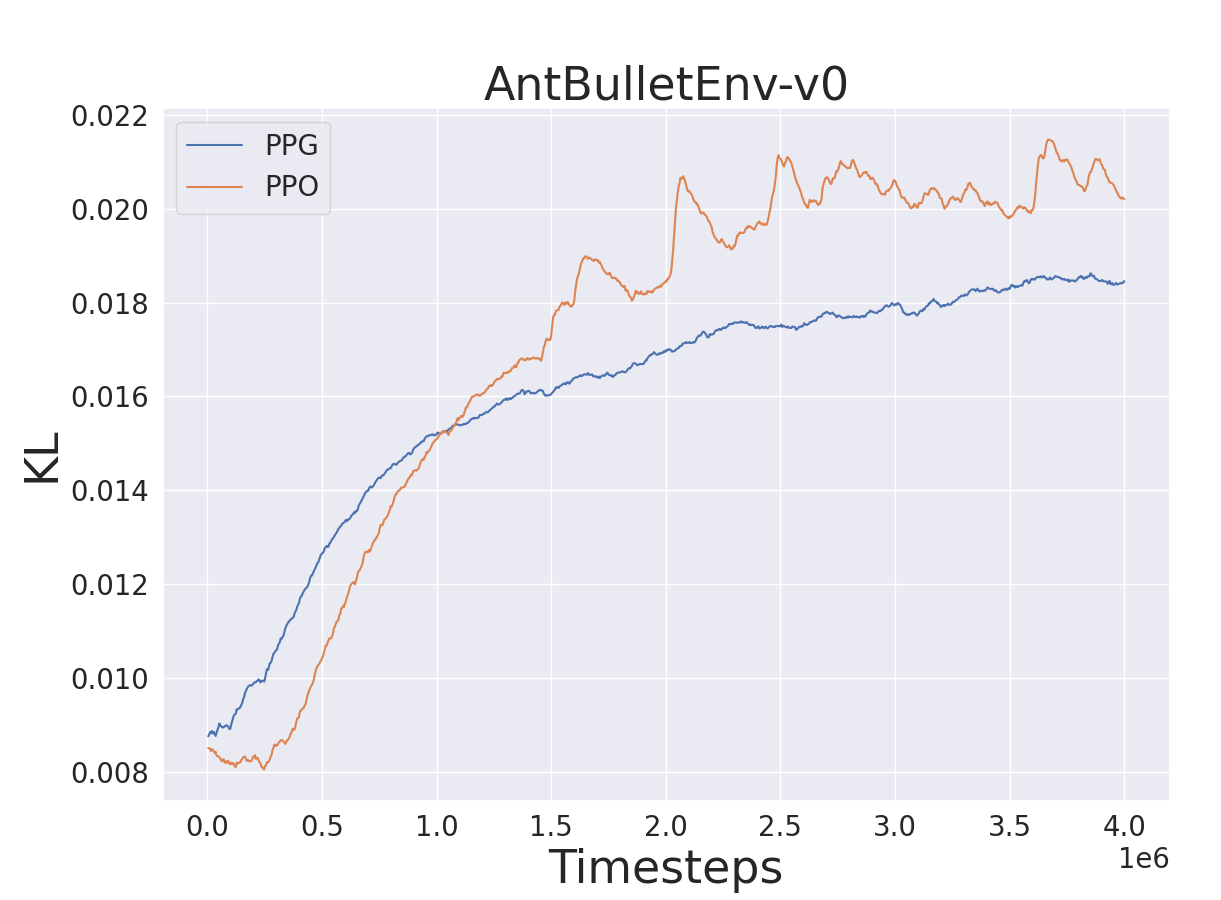} 
\includegraphics[width=0.5\columnwidth]{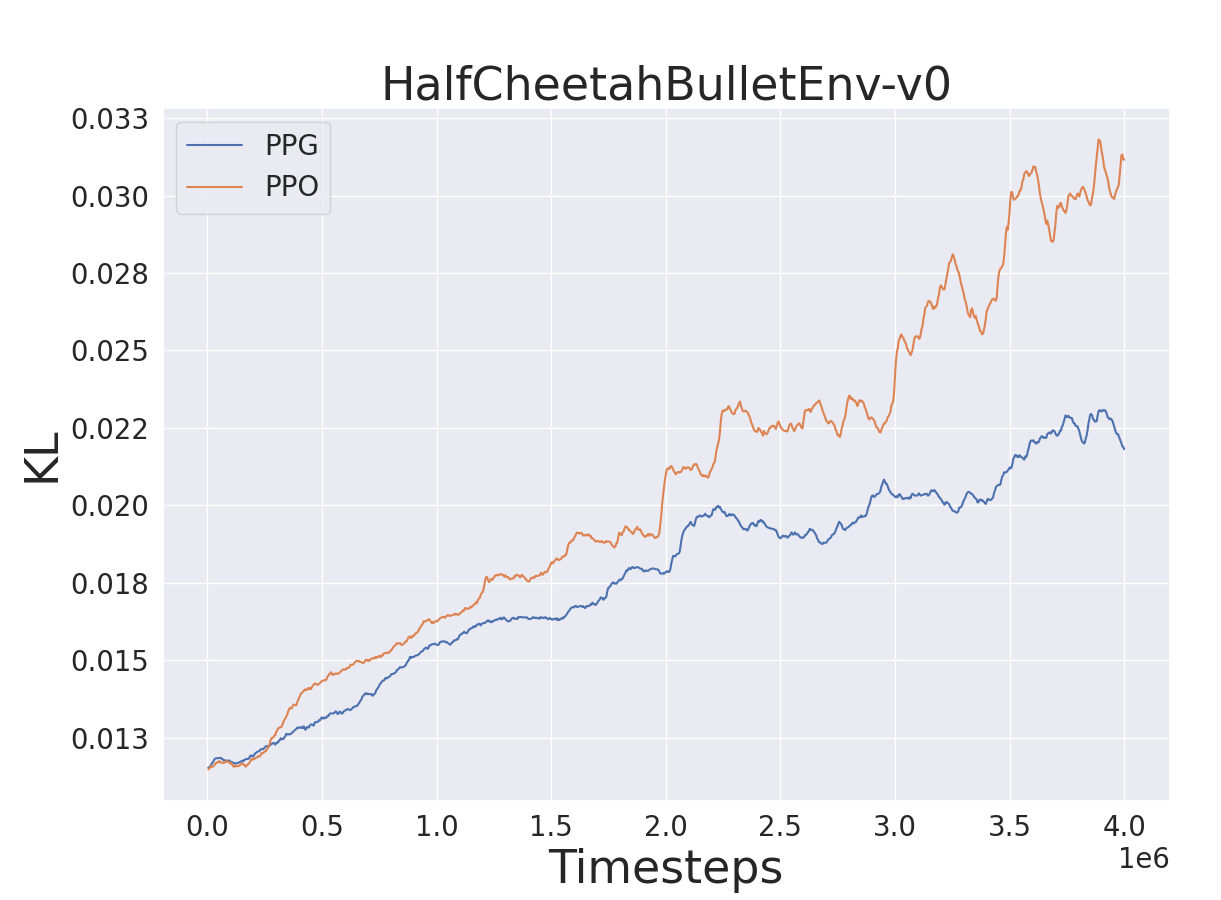}
\includegraphics[width=0.5\columnwidth]{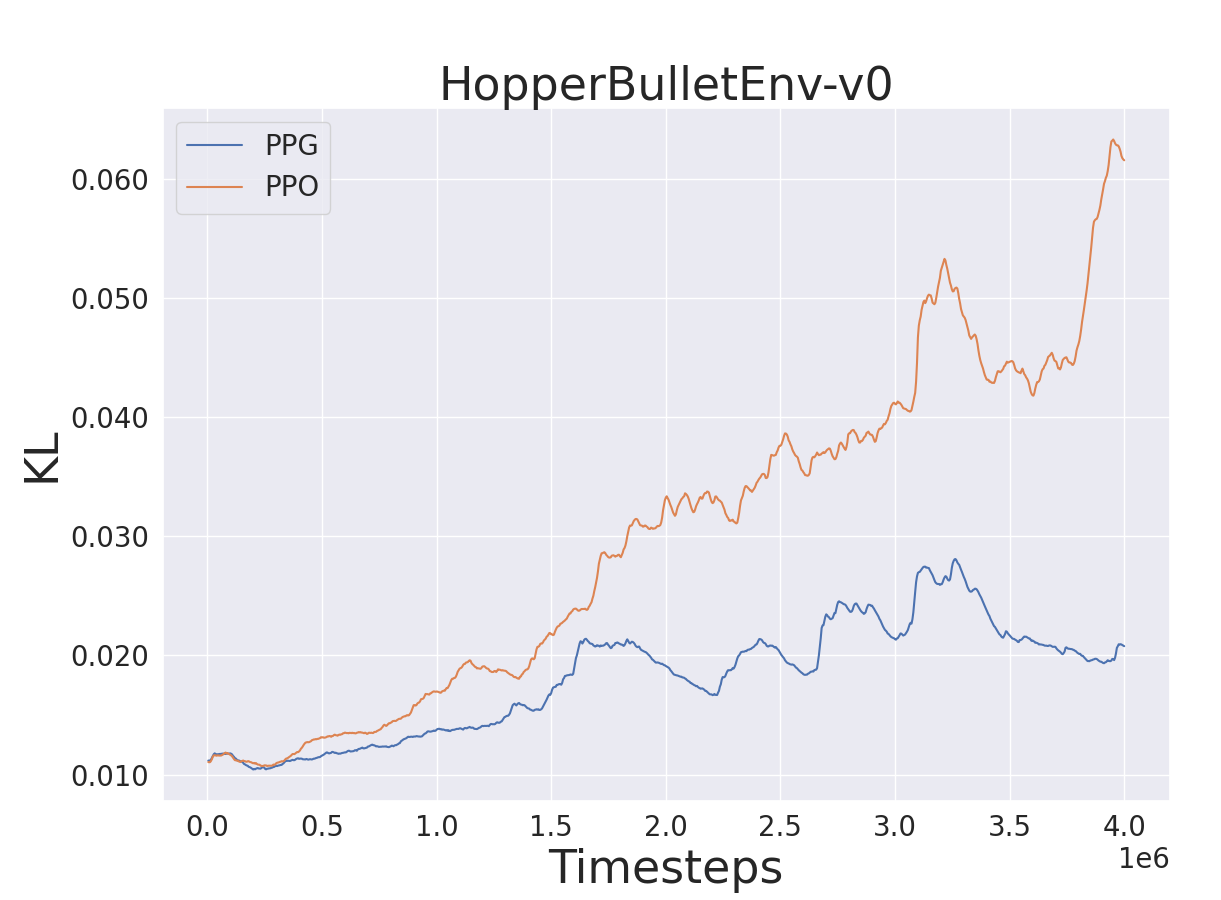}
\includegraphics[width=0.5\columnwidth]{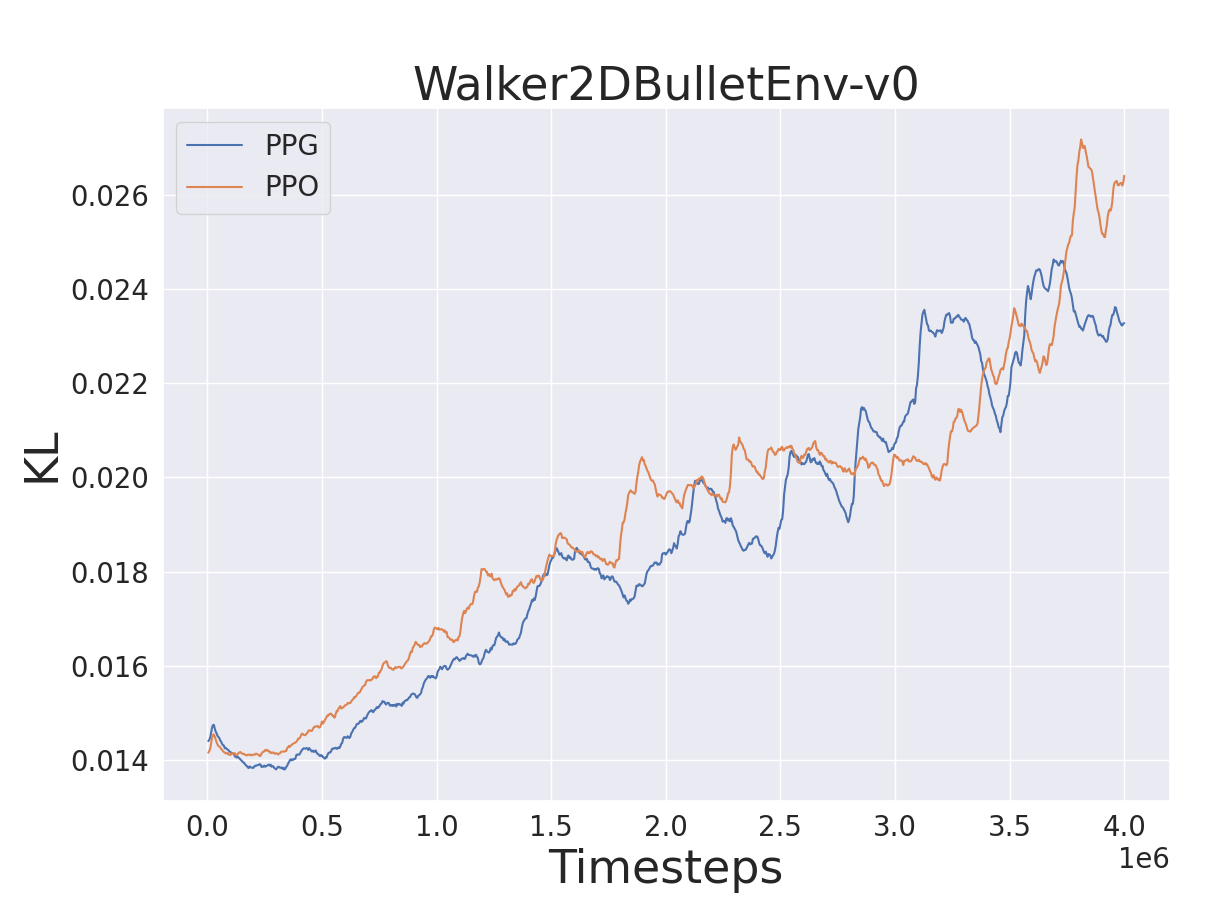}
\caption{KL divergences for PPG (light blue) and PPO (light orange). The graphs show the KL divergences of PPG tend to stay lower than that of PPO throughout the training.}
\label{fig:KL_Divergence_Experiments}
\end{figure*}

%------------------------------------------------------------------------
\begin{figure}[ht]
\centering
\begin{tikzpicture}

\draw[thin,->] (-3.5,0) -- (3.5,0) node [anchor=north] {$\hat A^t$};
\draw[thin,->] (0,-2.5) -- (0,2.5) node [anchor=south] {$\log\pi-\log\pi_{old}$};

\draw[very thin] (0, 1.0) node [anchor=east] {$u_b$} -- ( 3.5, 1.0);
\draw[very thin] (0,-1.0) node [anchor=west] {$l_b$} -- (-3.5,-1.0);
\fill [blue,opacity=0.1] (0., 1.0) rectangle ( 3.5, 2.5);
\fill [blue,opacity=0.1] (0.,-1.0) rectangle (-3.5,-2.5);

\draw[very thin,->] ( 1,-1.5) -- node[right]{\scriptsize Increase probability} ( 1,-1);
\draw (  0.2,-2.) node[right]{\scriptsize Actions with positive advantage};
\draw[very thin,->] (-1, 1.5) -- node[ left]{\scriptsize Decrease probability} (-1, 1);
\draw ( -3.5, 2.) node[right] {\scriptsize Actions with negative advantage};

\draw[very thin,->] ( 1, 0.3) -- node[right]{\scriptsize Increase probability} ( 1, 0.7);
\draw[very thin,->] (-1,-0.3) -- node[ left]{\scriptsize Decrease probability} (-1,-0.7);

\draw ( 2, 2.0) node {\scriptsize Desirable};
\draw (1.8, 1.6) node {\scriptsize (Probability is increased enough)};
\draw (-2.0,-1.8) node {\scriptsize Desirable};
\draw (-1.8,-2.2) node {\scriptsize (Probability is decreased enough)};

\end{tikzpicture}
\caption{Advantage-Policy plane. In left half plane, since $\hat A_t<0$, probability should decrease and on the right half plane, since $\hat A_t>0$, probability should increase. We apply a clipping such that gradient is zero in the blue shaded regions since log probabilities have updated more than the clipping thresholds $u_b$ and $l_b$. See Fig. \ref{fig_ppg_po} for an example.}
\label{fig_ap}
\end{figure}
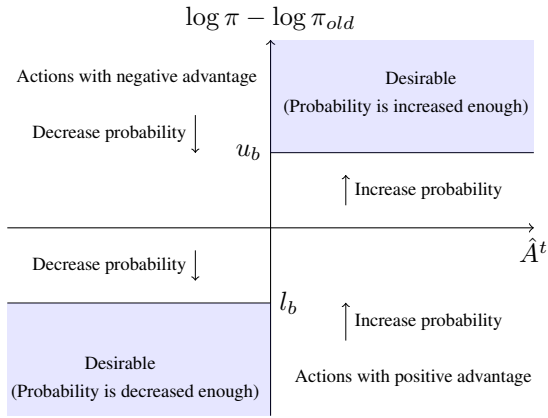

In a policy update iteration, if $\hat A_t>0$, the policy parameters should be updated to increase the probability $\pi_{\theta}(a_t|s_t)$, and if $\hat A_t<0$, the policy parameters should be updated in the direction of decreasing $\pi_{\theta}(a_t|s_t)$. As illustrated in Fig. \ref{fig_ap}, the gradient of $L^{\scalebox{0.55}{$PPG$}}$ should have a downward $y$-component in the $2^{nd}$ and the $3^{rd}$ quadrants, and a upward $y$-component in the $1^{st}$ and the $4^{th}$ quadrants.

Suppose a sample is in the $2^{nd}$ quadrant. This means that the probability has been increased even though the action is disadvantageous. To penalize this undesirable update, we do not set a clipping upper bound. Then all samples in the $2^{nd}$ quadrant are used in the gradient ascent, and their probabilities are likely to be decreased for the next policy update. Now, suppose that a sample is in the $4^{th}$ quadrant. This represents that the probability of that advantageous action is decreased. Therefore, the objective function should not clip the sample to penalize all of such cases.

Samples in the $1^{st}$ quadrant indicate that probabilities of advantageous actions are increased. These samples are updated properly unless the amount of log probability increase is too much. Therefore, we set an upper bound to prevent the policy from being too deterministic. Samples above this bound will have zero gradient, and hence are not forced to increase the probability further. This procedure also similarly works for samples in the $3^{rd}$ quadrant. The lower bound in the $3^{rd}$ quadrant prevents excessive probability decreases by forcing the gradients to be zero for all samples that the difference log probabilities is lower than the lower bound. Notice that clipped samples have zero gradient, so the samples in the desirable areas are effectively removed from the next policy update iteration. Thus the clipping makes the per iteration batch size small and only focuses on the samples that should be reflected in the next update.

In practice, as shown in Fig. \ref{fig_ppg_po}, at the beginning of the policy iteration, all samples are in the $x$ axis, and as iteration continues, samples start spreading into all the quadrants. Although a large number of samples exist in the $2^{nd}$ and the $4^{th}$ quadrants, the objective function (the red lines top far right image in Fig. \ref{fig_ppg_po}) indeed increases. A clip line ($u_b=0.2$) is sometimes observable as shown in the iteration 80. The iteration can be continued further until both of the clip lines are clearly observable, however, this typically hampers the performance.

Finally, we would like to note that one can design a variety of different clipping strategies such as slopped clip lines. In our tests, we obtained mixed results depending on environments. Thus, we left the exploration as a future work.

\begin{figure*}[t]
\centering
\includegraphics[width=0.499\columnwidth]{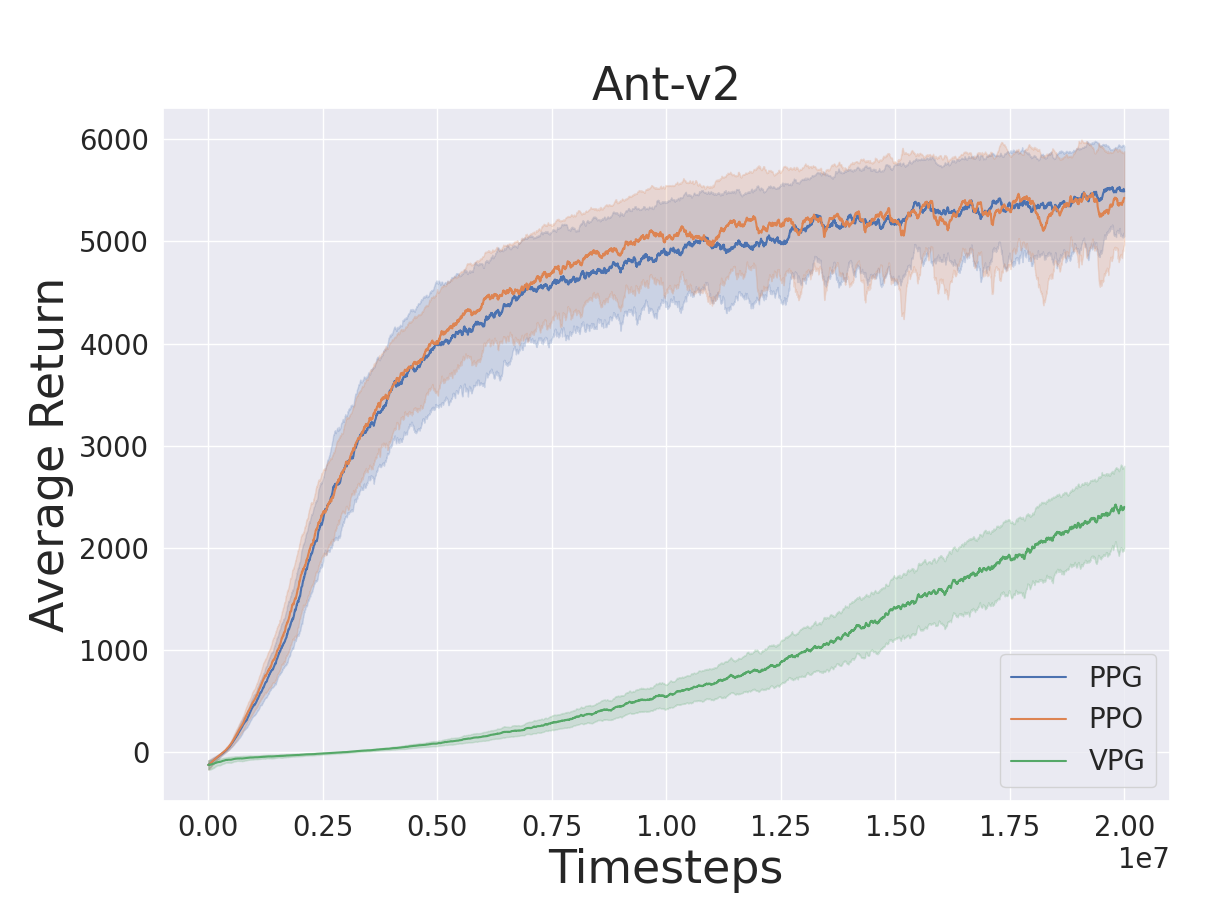} 
\includegraphics[width=0.499\columnwidth]{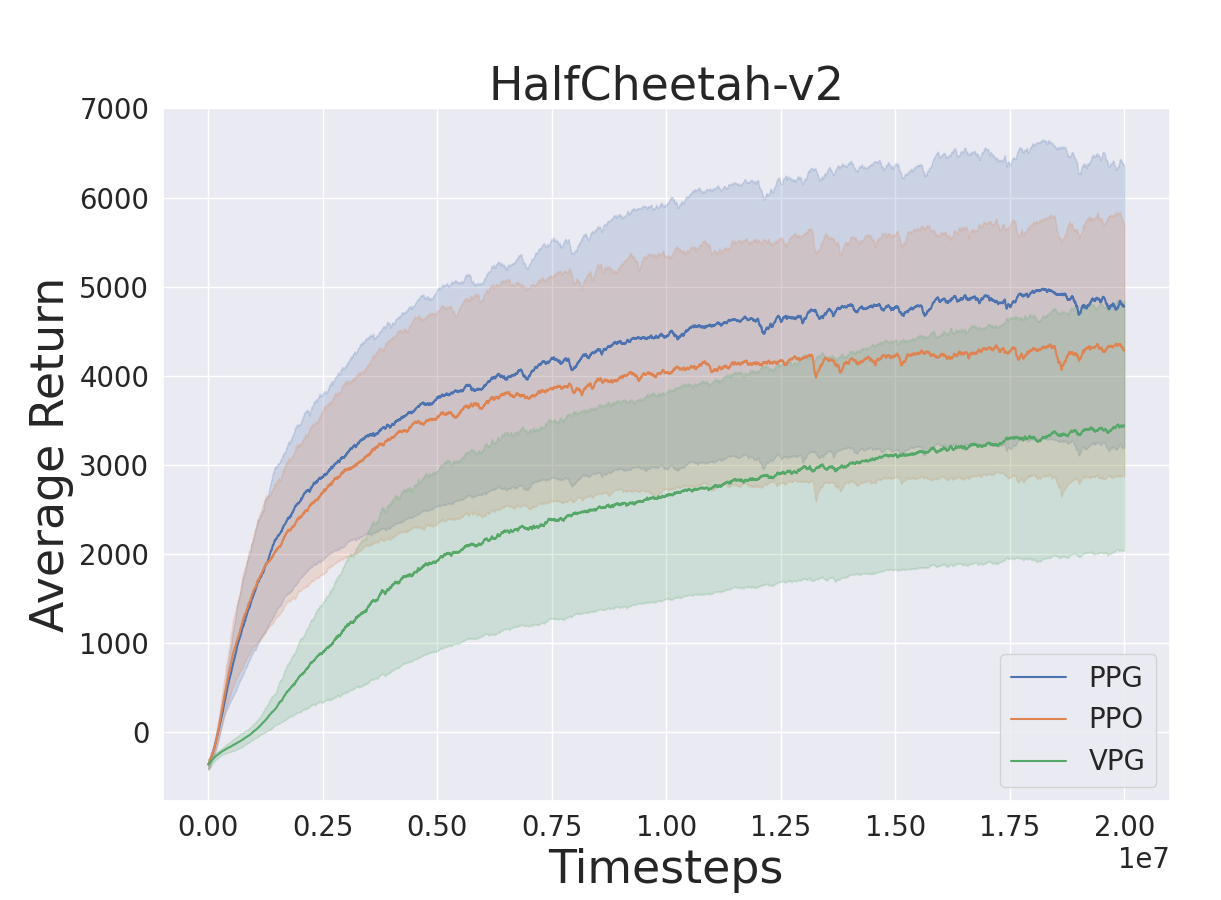}
\includegraphics[width=0.499\columnwidth]{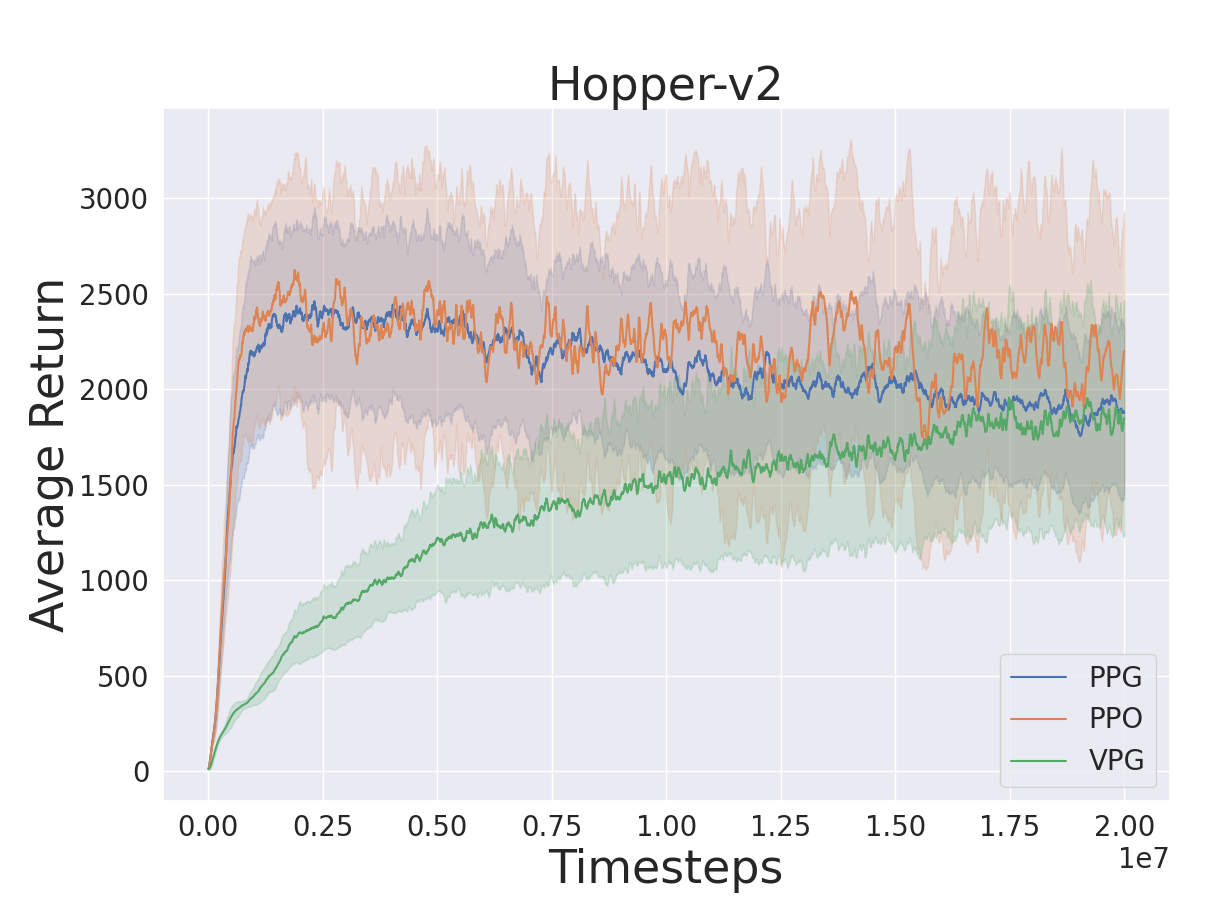}
\includegraphics[width=0.499\columnwidth]{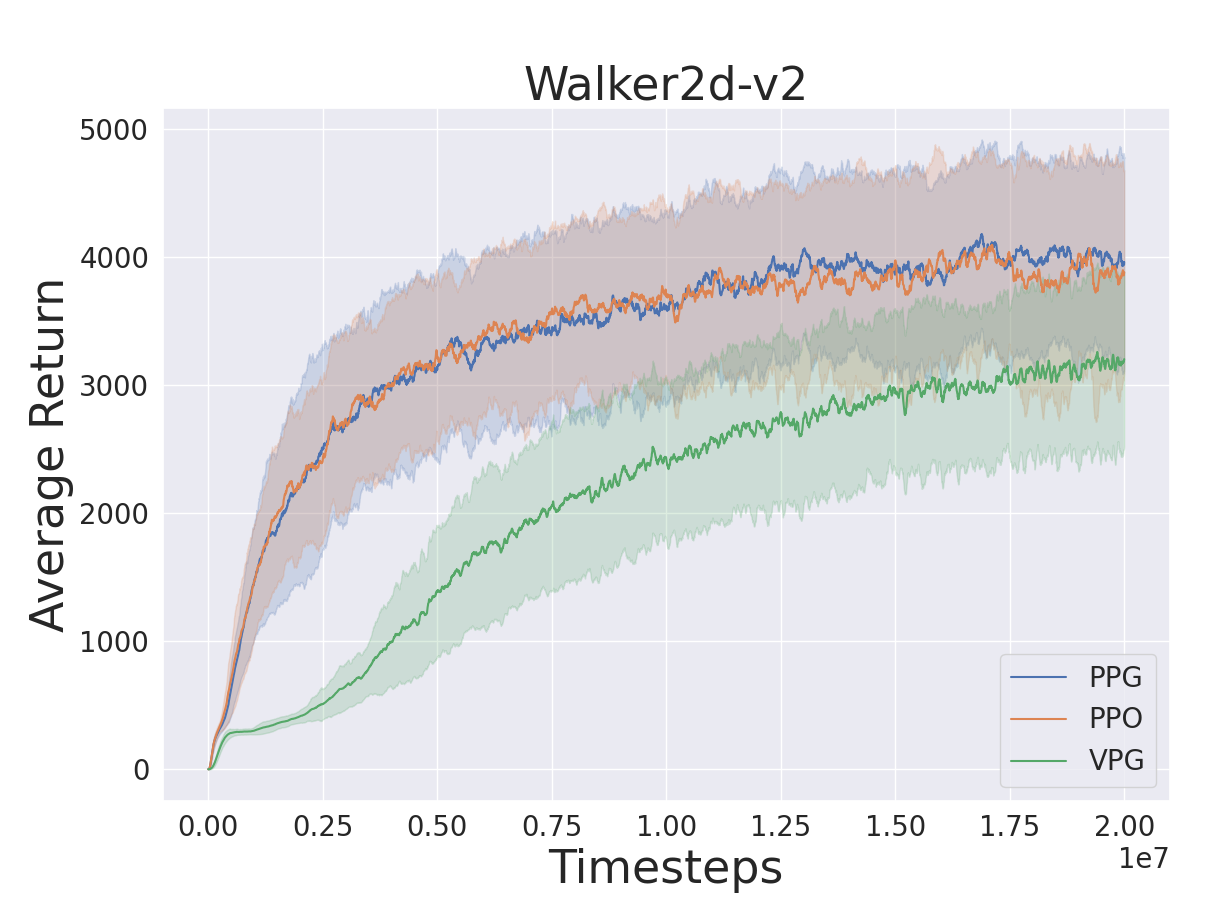}\\
\includegraphics[width=0.499\columnwidth]{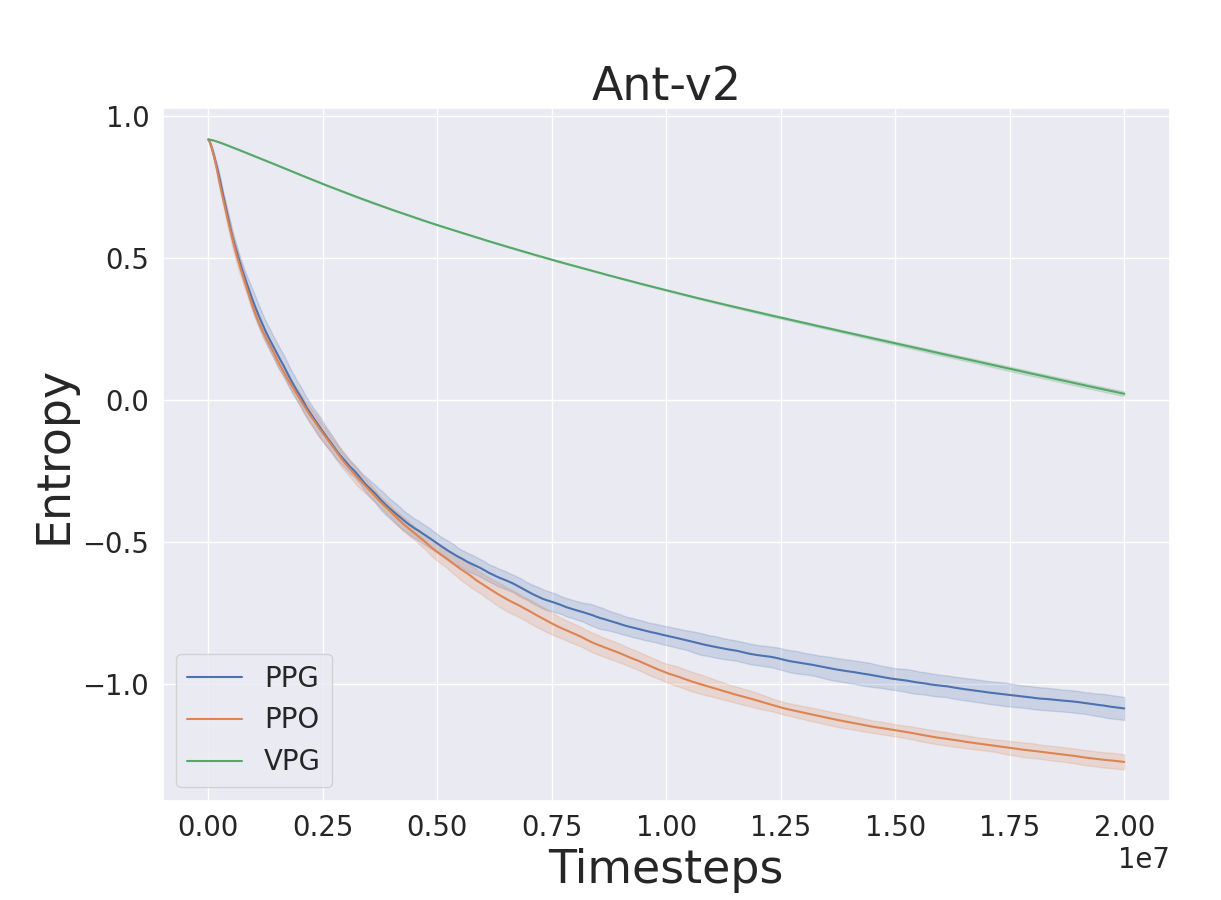}
\includegraphics[width=0.499\columnwidth]{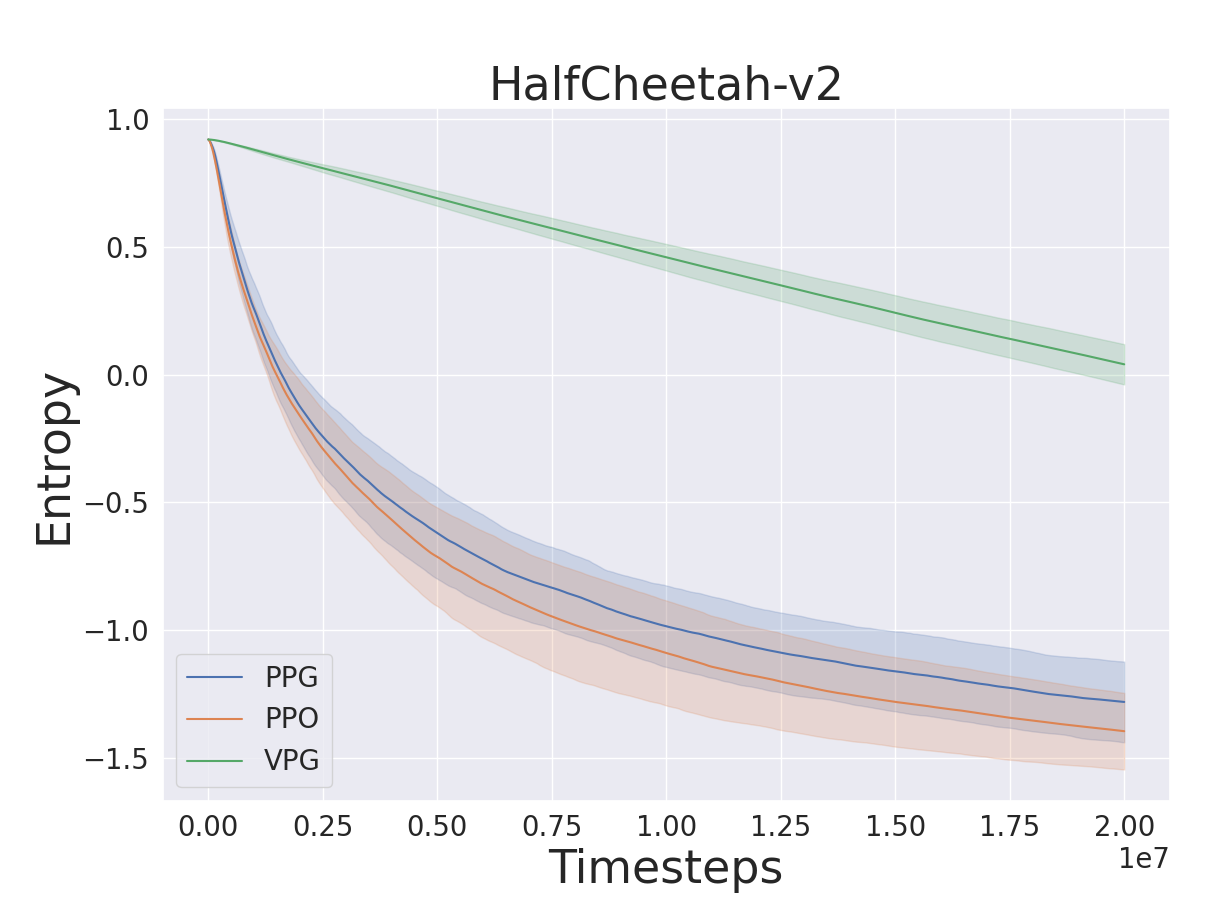}
\includegraphics[width=0.499\columnwidth]{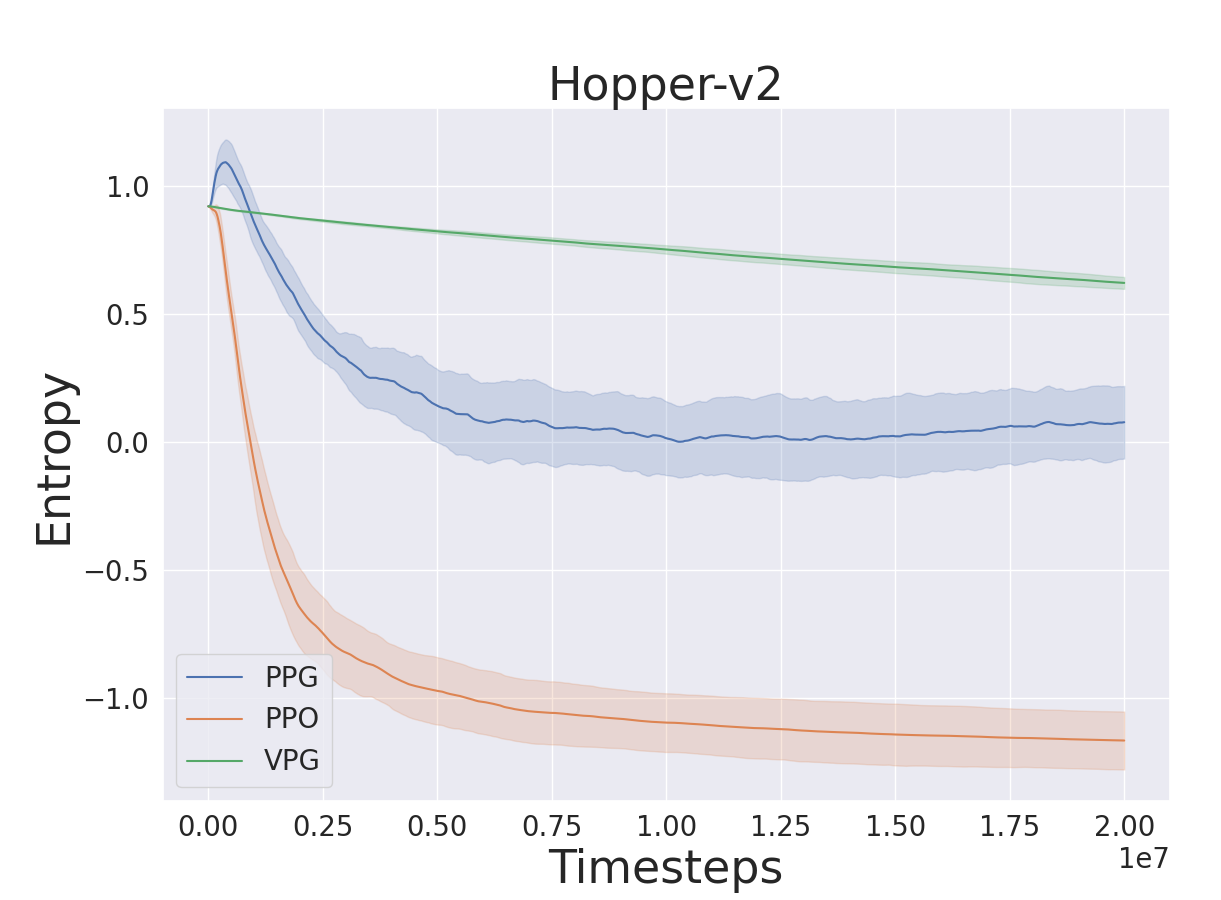}
\includegraphics[width=0.499\columnwidth]{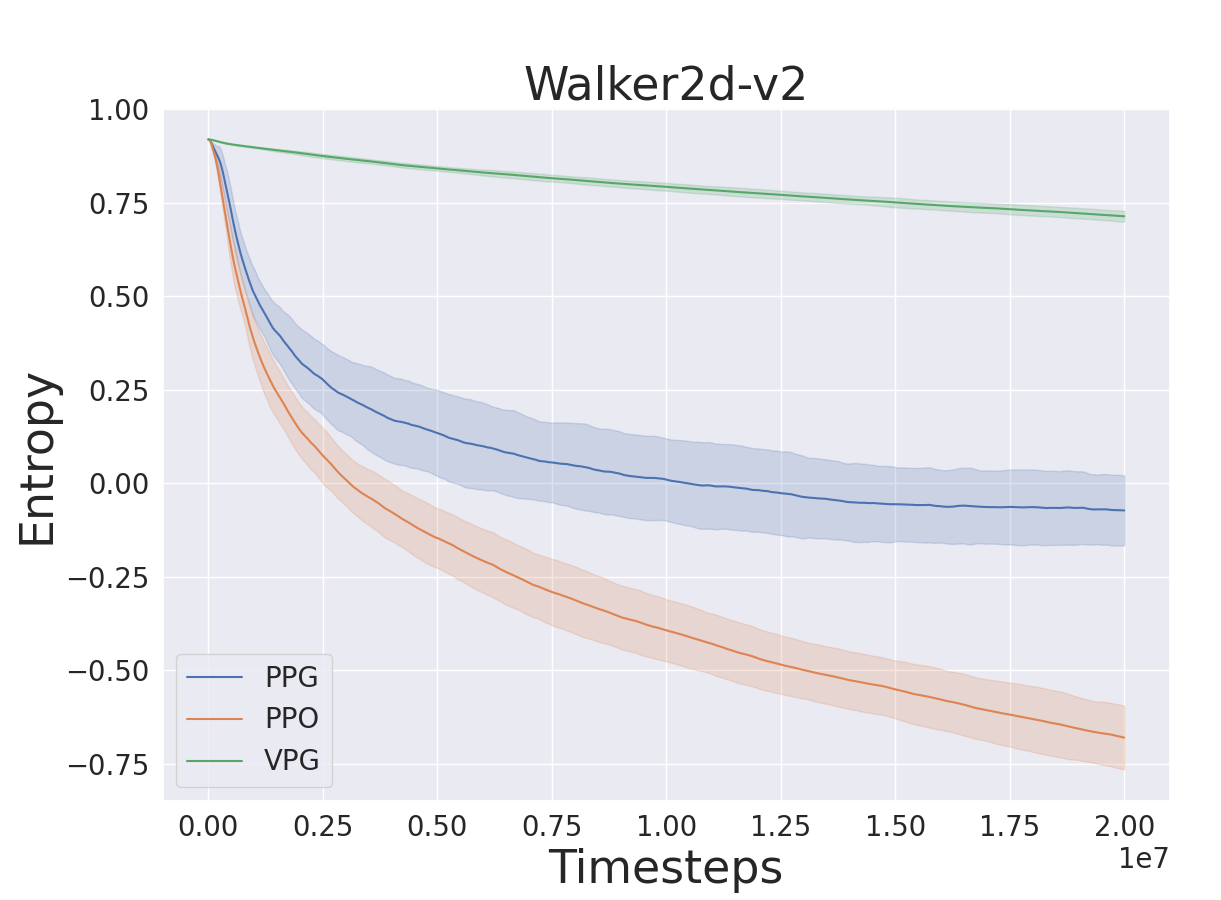}
\caption{Average return (top) and differential entropy (bottom) from 20M interactions with the four OpenAI Gym environments with the MuJoCo physics engine. PPG and PPO perform similarly. PPG tends to produce more stochastic solutions.}
\label{fig:MuJoCo_Result}
\end{figure*}

\begin{figure*}
\centering
\includegraphics[width=0.499\columnwidth]{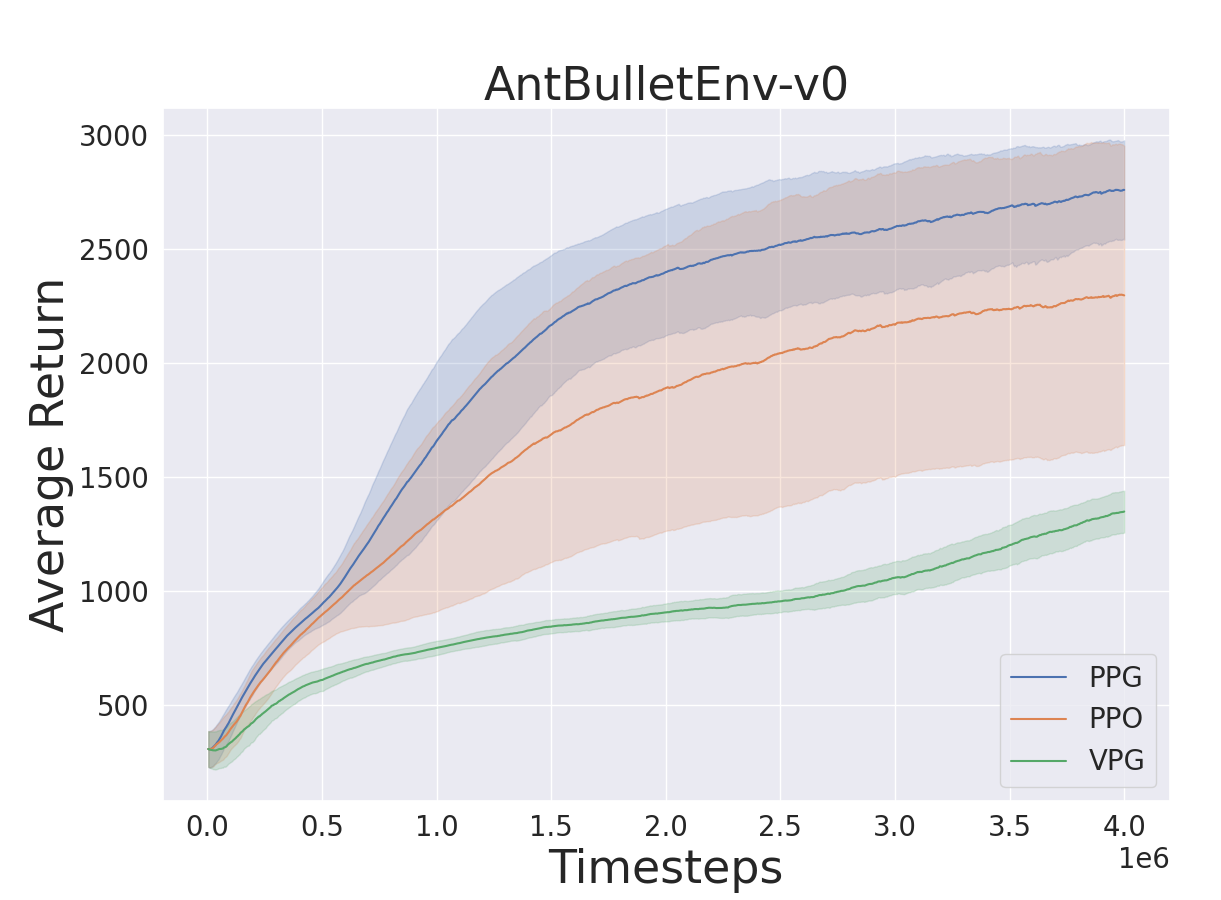} 
\includegraphics[width=0.499\columnwidth]{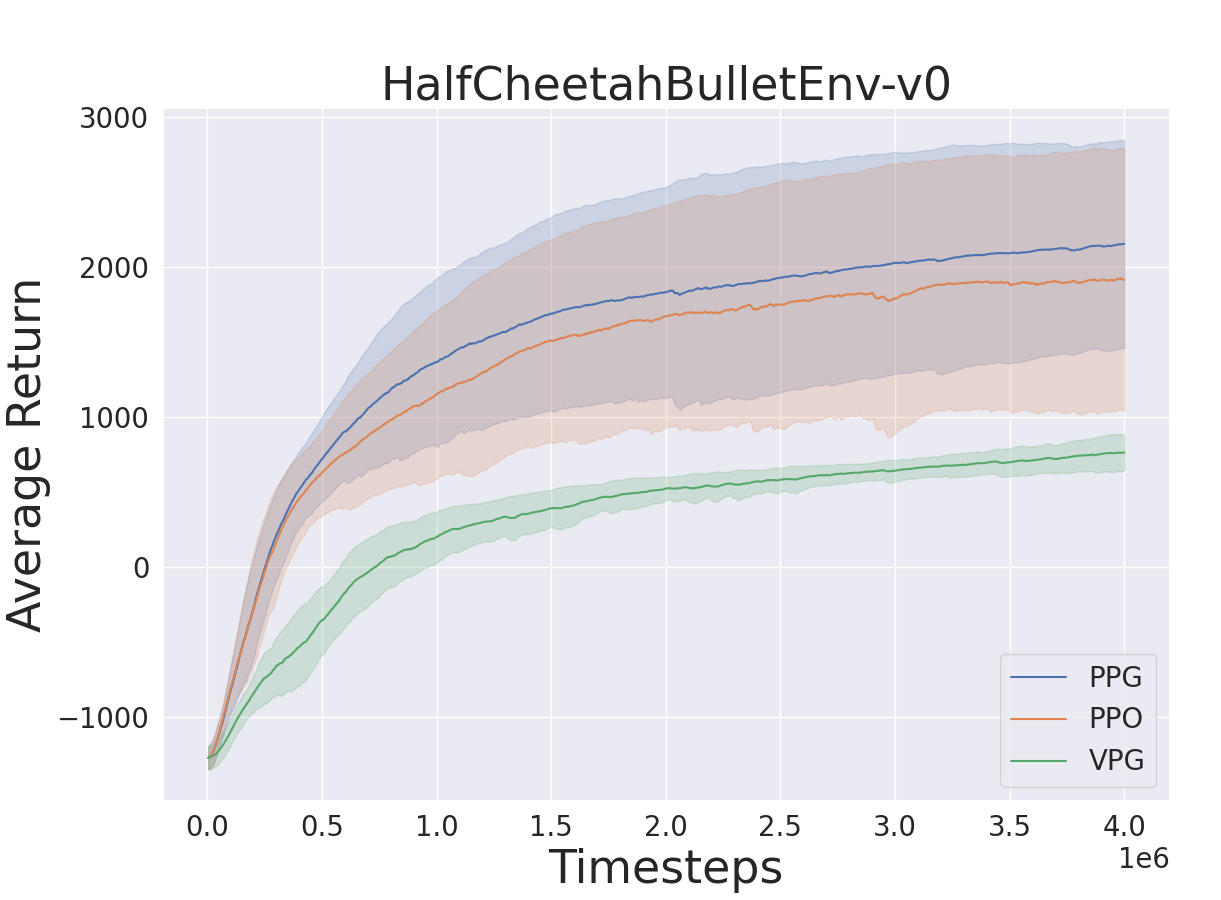}
\includegraphics[width=0.499\columnwidth]{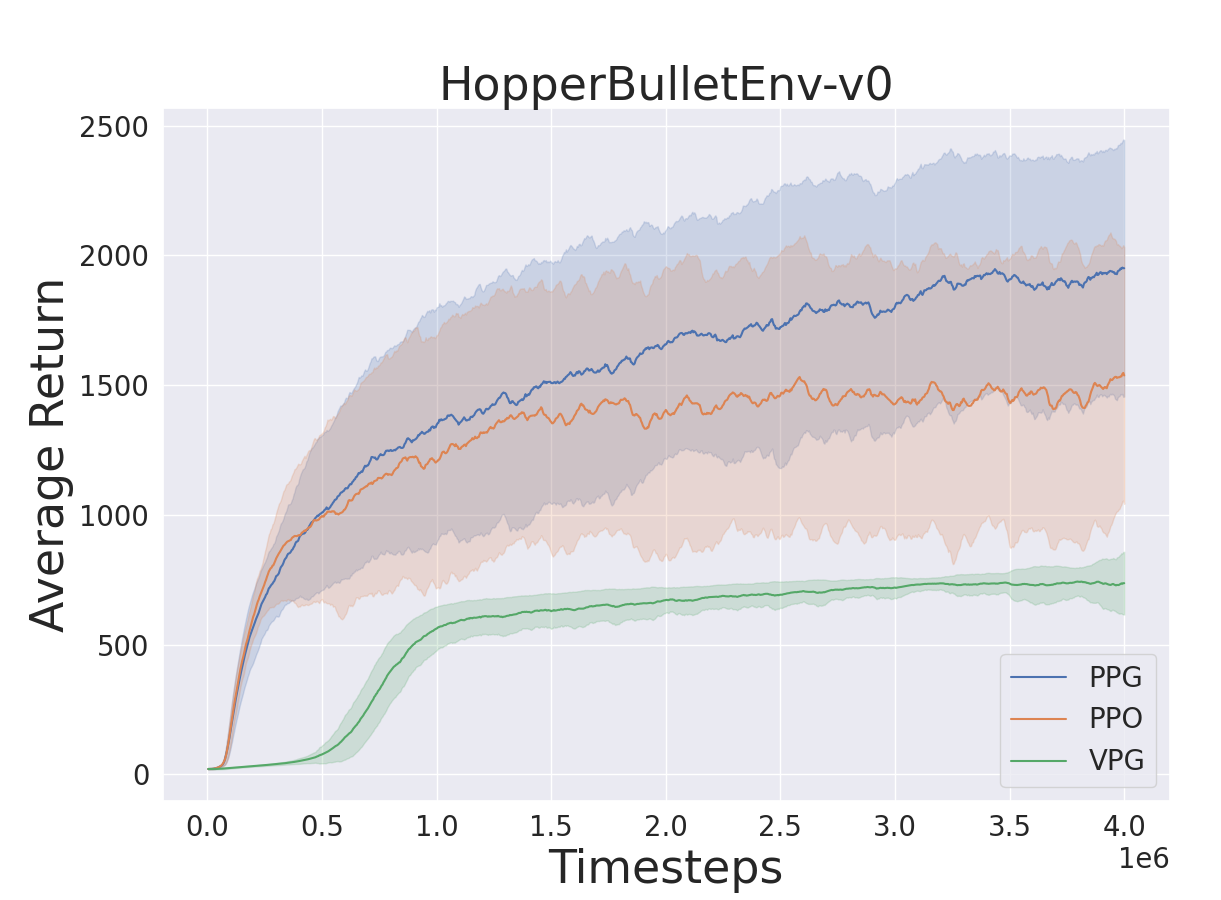}
\includegraphics[width=0.499\columnwidth]{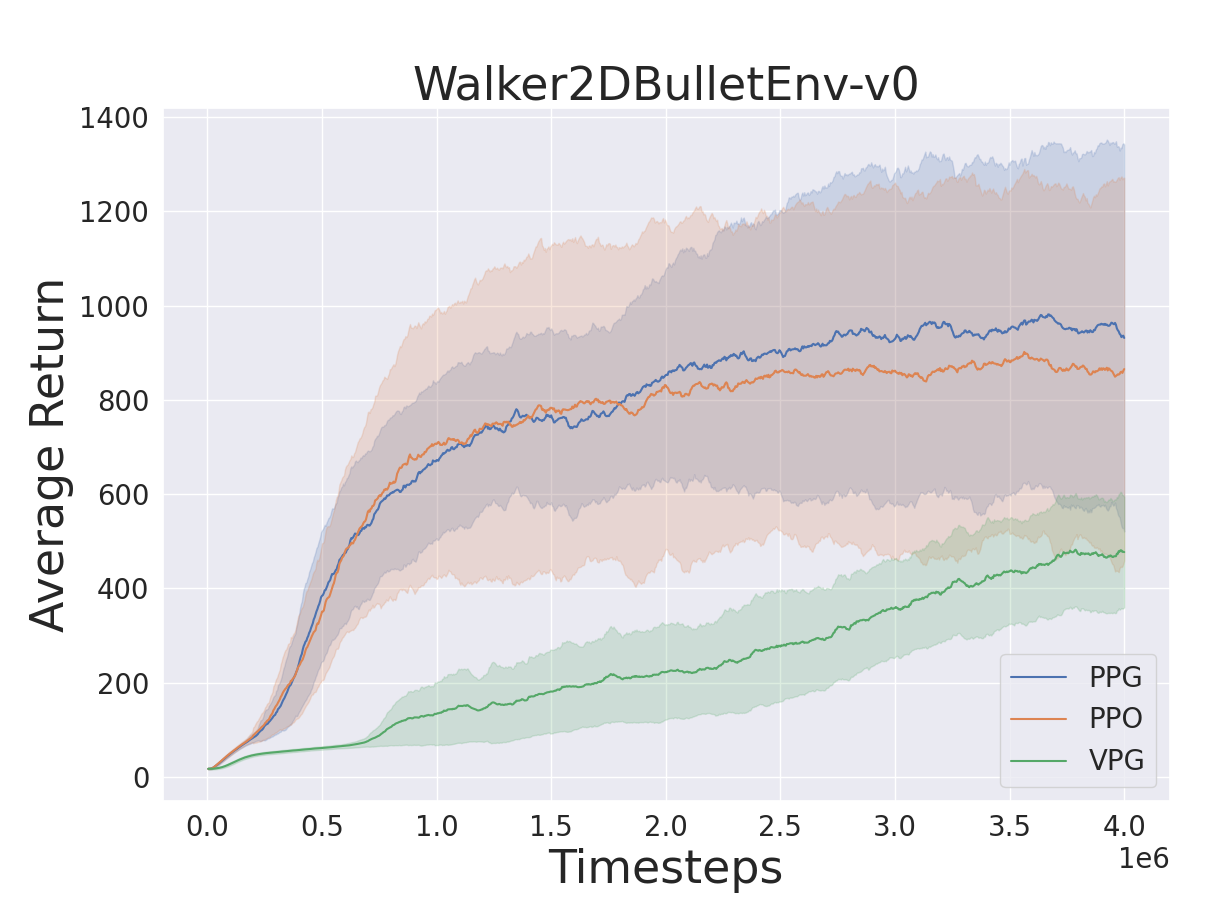}\\
\includegraphics[width=0.499\columnwidth]{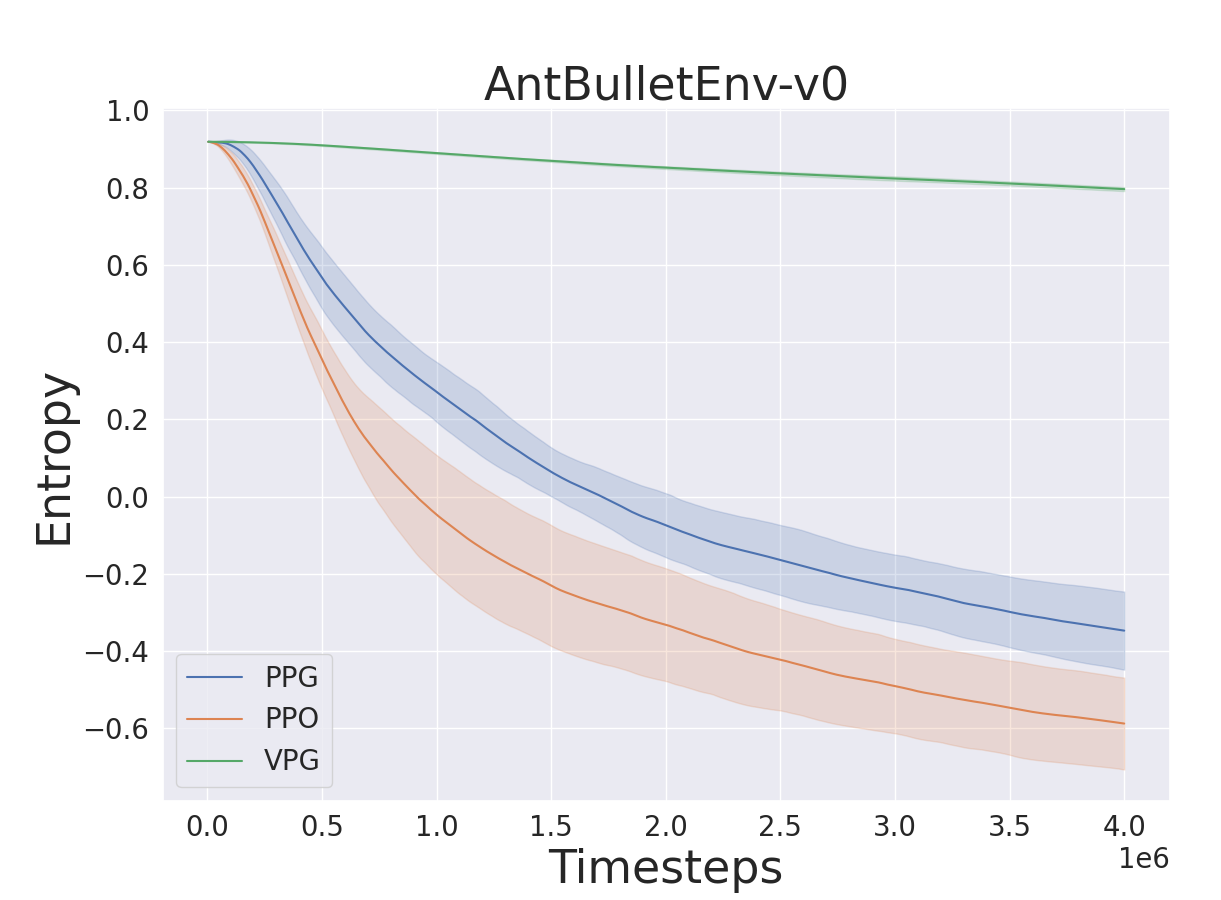}
\includegraphics[width=0.499\columnwidth]{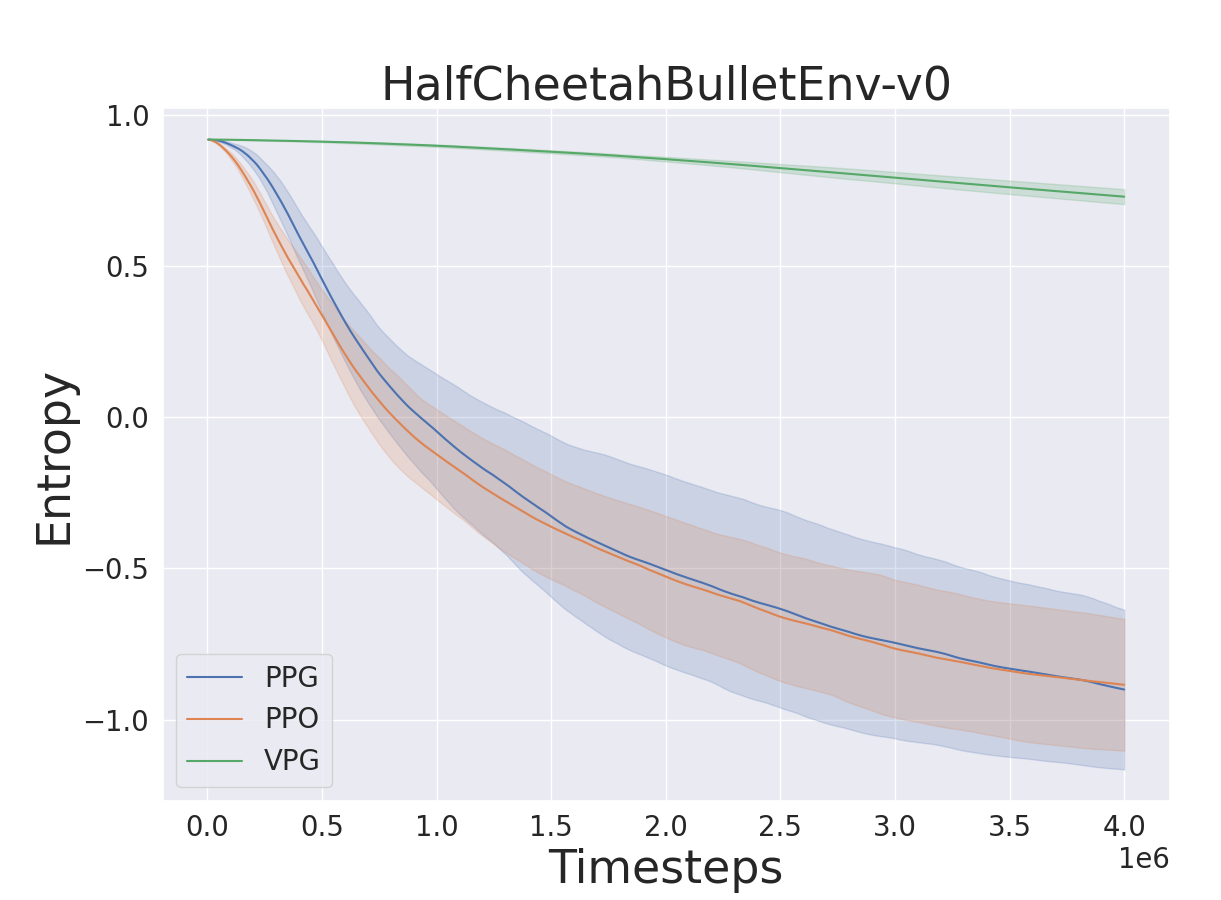}
\includegraphics[width=0.499\columnwidth]{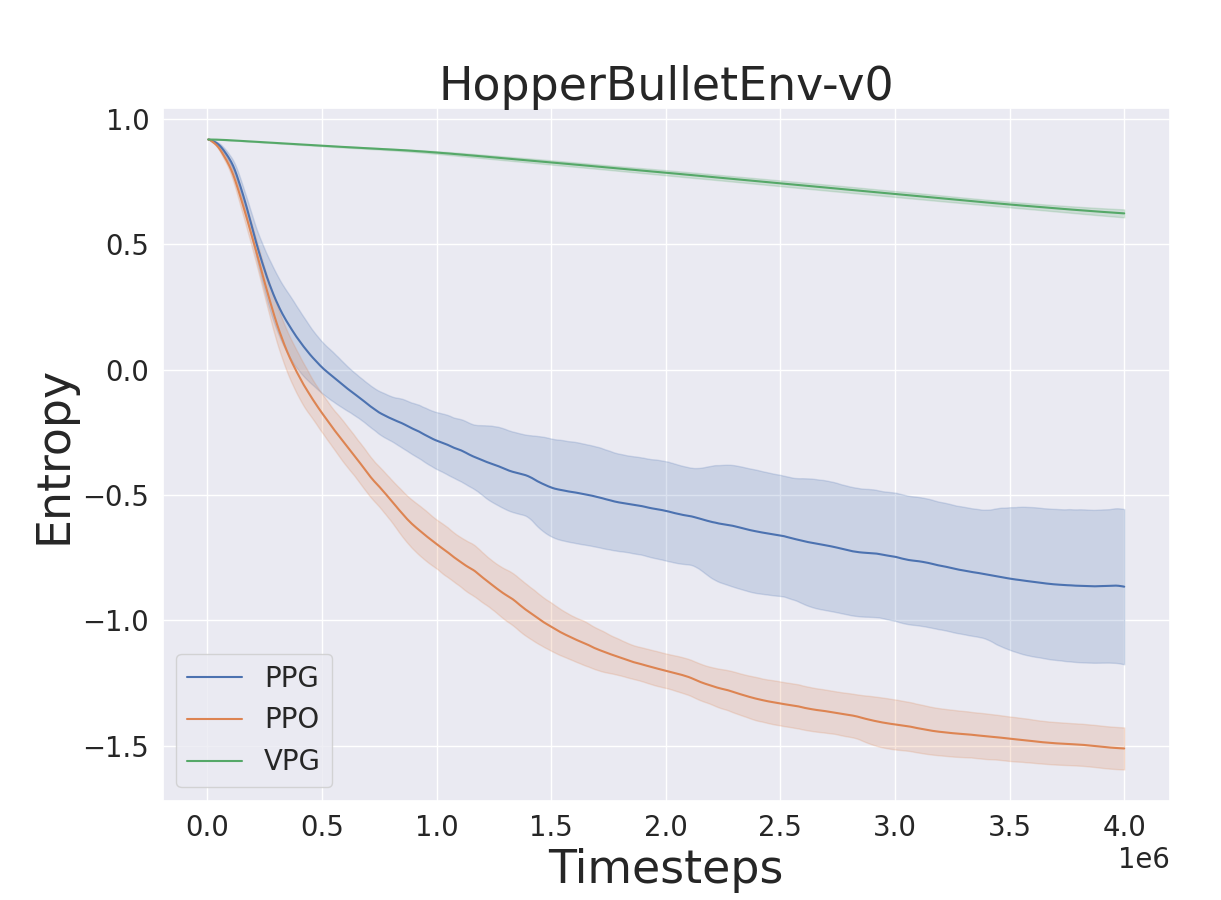}
\includegraphics[width=0.499\columnwidth]{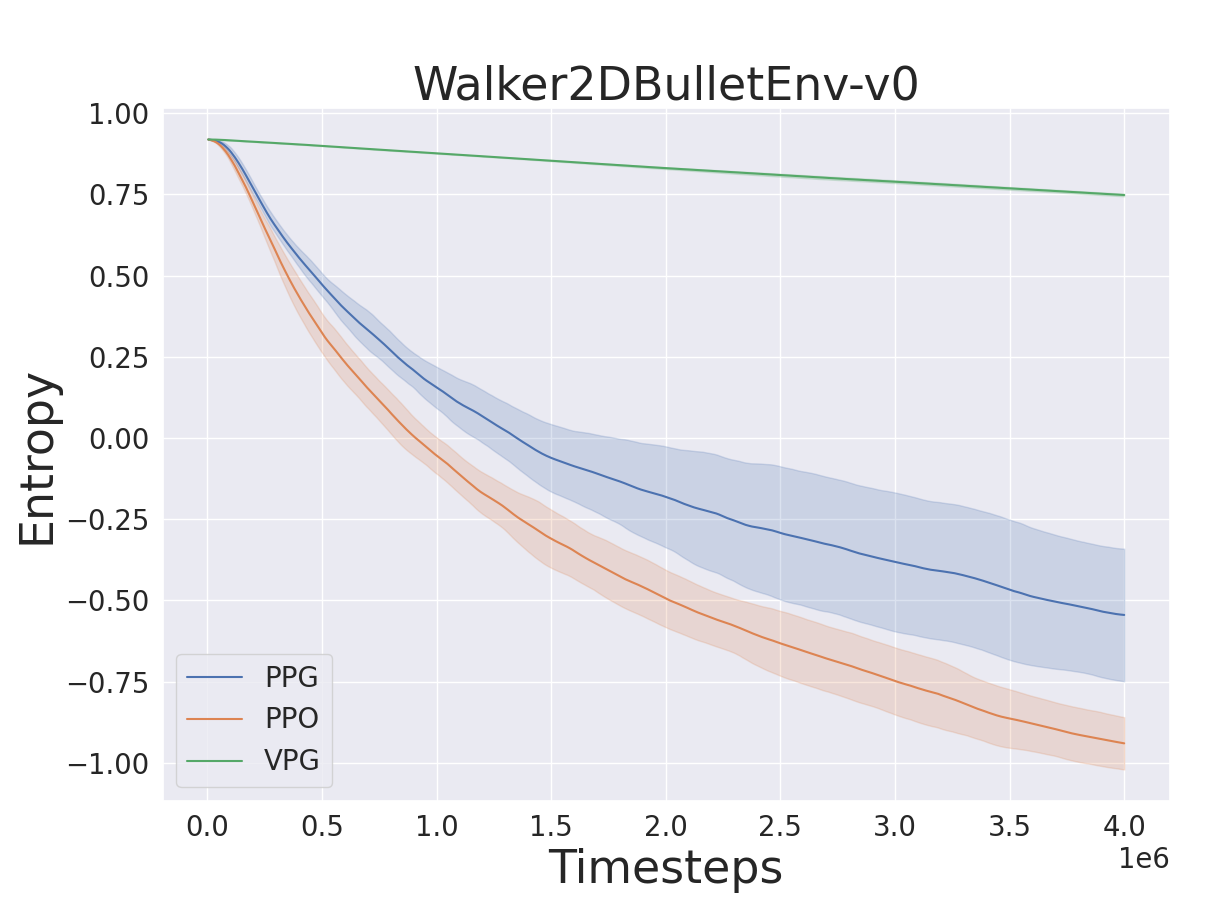}
\caption{Average return (top) and differential entropy (bottom) from 4M interactions with the four Bullet environments. PPG and PPO perform similarly. PPG tends to produce more stochastic solutions.}
\label{fig:Bullet_Result}
\end{figure*}

\section{Discussions on KL Divergence}
As a policy update iteration continues, $\pi_\theta$ becomes different from $\pi_{\theta_{old}}$ as demonstrated in Fig. \ref{fig_ppg_po}. The KL divergence, or relative entropy, measures the difference between $\pi_\theta$ and $\pi_{\theta_{old}}$ as
\begin{gather}
    D_{KL}(\pi_{\theta} || \pi_{\theta_{old}}) = \int \pi_{\theta}(a|s)\log\frac{\pi_{\theta}(a|s)}{\pi_{\theta_{old}}(a|s)} da
\label{eq:KL_Divergence}\end{gather}
$D_{KL}$ can be approximated by the following $D_{MC}$, i.e., $D_{KL} \to D_{MC}$ as $n \to \infty$ \cite{hershey2007approximating}.
\begin{gather}\begin{split}
    D_{MC}(\pi_{\theta} || \pi_{\theta_{old}}) 
    &= \frac{1}{\hat N}\sum_{i, t=1}^{N,T_i} \left(\log\pi_{\theta}(a|s) - \log\pi_{\theta_{old}}(a|s)\right)\\
    &= \frac{1}{\hat N}\sum_{i, t=1}^{N,T_i} d_{i,t}
\label{eq:Approximated_KL_Divergence}\end{split}\nonumber
\end{gather}
Therefore, the average $y$ coordinates of all samples in the advantage-policy plane is just the approximated KL divergence $D_{MC}$. The objective $L^{\scalebox{0.55}{$PPG$}}_{\scalebox{0.55}{$NCLIP$}} = \frac{1}{\hat N}\sum_{i, t=1}^{N,T_i} d_{i,t}(\theta) \hat A_t$ has some relation to $D_{MC}$. Although  $d_{i,t}(\theta)$ and $\hat A_t$ are correlated, one can easily see some trivial relations. For example, 
$L^{\scalebox{0.55}{$PPG$}}_{\scalebox{0.55}{$NCLIP$}} \le \max(\hat A_t)D_{MC}$ in the right half plane where $\hat A_t>0$. Since $\hat A_t$ is normalized to have the standard deviation of 1, $\max(\hat A_t)$ is not large, typically less than 10. 

When the iteration starts, all samples are on the x-axis. As the current policy is updated, the samples gradually spread over to all quadrants, while moving to the first and third quadrants are encouraged by the gradient. In the left half plane, since samples are driven to have decreased probabilities by the optimization, the contribution to the approximate KL divergence, $\sum_{\hat A_t<0} d_{i,t}$ is likely to be negative. In the right half plane, $\sum_{\hat A_t<0} d_{i,t}$ is likely to be positive. Since the policy network outputs log probabilities, the log probability changes in both of the half planes are often similar in magnitude. Therefore, the two KL contributions tend to cancel, yielding a small approximate KL divergence as shown in Fig. \ref{fig:KL_Divergence_Experiments}.

Finally, we note that we have performed preliminary tests using a number of $D_{KL}$ constraint ideas such as $D_{KL}$ penalty added to the loss $L^{\scalebox{0.55}{$PPG$}}$ with or without various dead zones designs, for example, penalize only if $d_{i,t}>c$, where $c$ is a small constant. We also penalized using the average of exact KL divergence at each sample. In general, we found $D_{KL}$ is easy to control in PPG iterations, and the results were often positive, yet a principle that guides the choice among these methods and a study on general applicability on various environments need more thorough investigations, and hence, are left as future works.

\section{Experiments}

\subsection{Algorithms and Settings}
We presents Alg. \ref{alg:Algorithm} based on \cite{SpinningUp2018, baselines}. The policy $\pi_{\theta}$ and the value network $V_{\phi}$, parameterized by $\phi$, are the fully connected networks with two layers of size 64. We used hyperbolic tangent activation functions for all layers. For each epoch, after extracting trajectory samples, $A_{t}^{GAE}$ is computed and normalized. For VPG, the policy iteration is done once, and for PPG and PPO, the iteration is continued up to 80.

PPG and PPO perform the clipping step to update the policy within the trust region. This clipping functions have the clip range (\ref{eq:Loss_PPG}), (\ref{eq:Loss_PPO}) (In our experiments, $u_b = 0.2,\;l_b = -0.2,\; \epsilon = 0.2$). The lines 9 to 11 in Alg. \ref{alg:Algorithm} implement a secondary optimization technique: if the approximated KL divergence reaches to the target $D_{\scriptsize MC}^{Target}$, which we choose 0.015, the policy update iteration is halted. Without this policy learning became unstable in most of the environments. After the policy iteration, $V_{\phi}$ is updated by minimizing the mean square error $||\hat R_t - V_\phi||^2$.

\begin{algorithm}
    \caption{VPG, PPO, PPG}
    \label{alg:Algorithm}
    \begin{algorithmic}[1]
    \STATE \textbf{Input:} an initialized policy $\pi_{\theta}$ and value function $V_{\phi}$
    \STATE \textbf{for} \text{$k = 1 $ to epoch}
        \begin{ALC@g}
            \STATE Set $\pi_{\theta_{old}} \gets \pi_{\theta}$
            \STATE \text{Collect trajectories $\{\tau_{1}, \tau_{2}, \tau_{3}, ...\}$ from $\pi_{\theta_{old}}$}
            \STATE \text{Compute return $\hat{R}_{t}$}
            \STATE \text{Compute each $A_{t}^{GAE}$ using $V_{\phi}$}
            \STATE \textbf{for} {\text{$i = 1$ to $n$}} \quad\text{($n$=1 when VPG)}
                \begin{ALC@g}
                    \STATE \text{Compute loss \eqref{eq:Loss_VPG}, \eqref{eq:Loss_PPO}, or \eqref{eq:Loss_PPG}}
                    \STATE \text{$m_{KL} \gets$ the mean of the approximated $D_{MC}$}
                    \STATE \textbf{if} \text{$m_{KL} > D_{\scriptsize MC}^{Target}$} \textbf{and not} \text{VPG}
                    \begin{ALC@g}    
                        \STATE \text{\;\;\quad break} 
                    \end{ALC@g}
                    \STATE \textbf{end if}
                    \STATE \text{Update $\pi_{\theta}$ by maximizing loss \eqref{eq:Loss_VPG}, \eqref{eq:Loss_PPO}, or \eqref{eq:Loss_PPG}}
                \end{ALC@g}
            \STATE \textbf{end for}
        \STATE Update $V_\phi$ by minimizing $||\hat R_t - V_\phi||^2$
        \end{ALC@g}
    \STATE \textbf{end for}
  \end{algorithmic}
\end{algorithm}

\subsection{Evaluation on Benchmark Tasks}
We evaluated PPG on the OpenAI Gym robotics tasks \cite{DBLP:journals/corr/BrockmanCPSSTZ16}, and the corresponding the Bullet tasks (Ant-v2, HalfCheetah-v2, Hopper-v2, Walker2d-v2, AntBulletEnv-v0, HalfCheetahBulletEnv-v0, HopperBulletEnv-v0, Walker2DBulletEnv-v0). For each environment, we trained 10 models using ten seeds (10000,10001,10002,...,10009). In the OpenAI Gym tasks, the models are trained for 20 million interactions with the  each environments, and 4 million interactions for Bullet tasks. Training results are shown in Fig. \ref{fig:MuJoCo_Result}, \ref{fig:Bullet_Result}, and trained model evaluations are provided in Table. \ref{tb:MuJoCo_Bullet}. For this evaluation, we simulated 100 episodes for each model to measure the performance. 
We can observe that the entropy decays slower in PPG, while the average return is similar to PPO. In addition, in Fig. \ref{fig:KL_Divergence_Experiments}, PPG has smaller the KL divergence than PPO, under the exactly same condition in Alg. \ref{alg:Algorithm}.

\section{Conclusions}
We have shown that the surrogate objective can be replaced by a partial variation of the original policy gradient objective that has exactly the same gradient with the original policy gradient theorem based algorithm, VPG. Similar to PPO's clipping mechanism that updates a policy within the trust region, PPG  formulates the clipping in the log space. PPG shows the average returns comparable to PPO, the often higher entropy than PPO, and the smaller KL divergence changes than PPO.

\begin{table}[t]
\small
    \centering
    \caption{Trained model evaluations. Each entry has ten different trained models for 10 seeds. The average rewards are computed from 1000 episodes (100 episodes for each of 10 seeds). The entropy is the average of 10 models.}
    \begin{tabular}{lccc}
        \toprule
        \multicolumn{4}{c}{\textbf{OpenAI Gym (-v2)}} \\
        \toprule

        \textbf{(Reward)}&PPG&PPO&VPG\\
        \midrule
        Ant & 5527$\pm$947.5 & 5583$\pm$652.8 &2360$\pm$900.6 \\
        HalfCheetah & 4783$\pm$1708 & 4310$\pm$1537 & 3463$\pm$1480 \\
        Hopper & 1865$\pm$769.7 & 2292$\pm$970.2 &1756$\pm$871.9 \\
        Walker2d & 3943$\pm$1400 & 3957$\pm$1330 &3209$\pm$1307 \\
        \toprule
        \textbf{(Entropy)}&PPG&PPO&VPG\\
        \midrule
        Ant &  -1.08$\pm$0.04 & -1.27$\pm$0.02 & 0.02$\pm$0.01 \\
        HalfCheetah & -1.28$\pm$0.15 & -1.39$\pm$0.14 & 0.03$\pm$0.07 \\
        Hopper & 0.07$\pm$0.14 & -1.16$\pm$0.11 & 0.61$\pm$0.02 \\
        Walker2d & -0.07$\pm$0.09 & -0.68$\pm$0.08 & 0.71$\pm$0.01 \\
        \toprule
        \multicolumn{4}{c}{\textbf{}} \\
        \toprule
        \multicolumn{4}{c}{\textbf{Bullet (BulletEnv-v0)}} \\
        \toprule
        \textbf{(Reward)}&PPG&PPO&VPG\\
        \midrule
        Ant &  2763$\pm$300.8 & 2322$\pm$680.1 & 1363$\pm$146.6 \\
        HalfCheetah & 2148$\pm$729.0 & 1887$\pm$958.6 & 768$\pm$187.2 \\
        Hopper & 1850$\pm$780.0 & 1639$\pm$654.7 & 728$\pm$227.8 \\
        Walker2d &  958$\pm$522.2 & 873$\pm$515.8 & 482$\pm$292.9 \\
        \toprule
        \textbf{(Entropy)}&PPG&PPO&VPG\\
        \midrule
        Ant &  -0.35$\pm$0.10 & -0.59$\pm$0.11 & 0.79$\pm$0.01 \\
        HalfCheetah & -0.91$\pm$0.25 & -0.89$\pm$0.21 & 0.72$\pm$0.02 \\
        Hopper & -0.87$\pm$0.31 & -1.51$\pm$0.08 & 0.61$\pm$0.01 \\
        Walker2d & -0.55$\pm$0.20 & -0.94$\pm$0.07 & 0.74$\pm$0.01 \\
        \bottomrule
    \end{tabular}
    \label{tb:MuJoCo_Bullet}
\end{table}

\bibliography{references}
%\nocite{*}

\end{document}